\documentclass[12pt,a4paper]{article}

\usepackage[latin1]{inputenc}
\usepackage{url}
\usepackage{latexsym}
\usepackage{amstext}
\usepackage{algorithmic}
\usepackage{amsmath}
\usepackage{amssymb}
\usepackage{bbm}
\usepackage{mathrsfs}
\usepackage{setspace}

\newcommand{\mboxit}[1]{\mbox{\textit{#1}}}

\newcommand{\mboxsc}[1]{\mbox{\textsc{#1}}}

\newcommand{\comment}[1]{}
\newcommand{\ie}{\mbox{i.e.}}
\newcommand{\eg}{\mbox{e.g.}}
\newcommand{\cf}{\mbox{cf.}}
\newcommand{\viz}{\mbox{viz.}}
\newcommand{\wrt}{\mbox{w.r.t.}}

\newcommand{\etal}{\mboxit{et al.}}
\newcommand{\qed}{\rule{2mm}{2mm}}
\newcommand{\tuple}[1]{\langle #1 \rangle}
\newcommand{\defined}{\ensuremath{=_{\text{Def}}}}
\newcommand{\Set}{\ensuremath{\text{\it A}}}

\newcommand{\act}{\ensuremath{\text{\it a}}} 
\newcommand{\prp}{\ensuremath{\text{\it p}}} 
\newcommand{\lit}{\ensuremath{\text{\it l}}} 
\newcommand{\fml}{\ensuremath{\varphi}} 
\newcommand{\modfml}{\ensuremath{\varPhi}} 
\newcommand{\clause}{\ensuremath{\chi}} 

\newcommand{\Act}{\ensuremath{\mathfrak{Act}}}
\newcommand{\Prp}{\ensuremath{\mathfrak{Prop}}}
\newcommand{\Lit}{\ensuremath{\mathfrak{Lit}}}
\newcommand{\Fml}{\ensuremath{\mathfrak{Fml}}}

\newcommand{\atm}[1]{\ensuremath{\text{\it atm}(#1)}} 
 
\newcommand{\indep}[2]{\mathit{indep_{#1}}(#2)}

\newcommand{\et}{\wedge}
\newcommand{\ou}{\vee}
\newcommand{\imp}{\rightarrow}
\newcommand{\sii}{\leftrightarrow}

\newcommand{\Worlds}{\ensuremath{\text{\it W}}}
\newcommand{\AccRel}{\ensuremath{\text{\it R}}}
\newcommand{\Val}{\ensuremath{\text{\it V}}}
\newcommand{\ract}{\AccRel_{\act}}

\newcommand{\val}{\ensuremath{\text{\it val}}} 
\newcommand{\model}{\mathscr{M}}
\newcommand{\modelclass}[1]{\ensuremath{\mathcal{M}}_{#1}} 

\newcommand{\poss}[1]{\langle #1 \rangle}
\newcommand{\nec}[1]{[ #1 ]}
\newcommand{\antec}{\ensuremath{\varphi}}  
\newcommand{\conseq}{\ensuremath{\psi}} 

\newcommand{\Loaded}{\ensuremath{\text{\it loaded}}}
\newcommand{\Alive}{\ensuremath{\text{\it alive}}}
\newcommand{\Awake}{\ensuremath{\text{\it awake}}}
\newcommand{\Dead}{\ensuremath{\text{\it dead}}}
\newcommand{\Walking}{\ensuremath{\text{\it walking}}}

\newcommand{\HasGun}{\ensuremath{\text{\it hasGun}}}
\newcommand{\wait}{\ensuremath{\text{\it wait}}}
\newcommand{\load}{\ensuremath{\text{\it load}}}
\newcommand{\shoot}{\ensuremath{\text{\it shoot}}}
\newcommand{\tease}{\ensuremath{\text{\it tease}}}

\newcommand{\drop}{\ensuremath{\text{\it drop}}}
\newcommand{\Black}{\ensuremath{\text{\it black}}}
\newcommand{\White}{\ensuremath{\text{\it white}}}

\newcommand{\toggle}{\ensuremath{\text{\it toggle}}}
\newcommand{\On}{\ensuremath{\text{\it on}}}
\newcommand{\Off}{\ensuremath{\text{\it off}}}

\newcommand{\Saved}{\ensuremath{\text{\it saved}}}
\newcommand{\Mbox}{\ensuremath{\text{\it mbox}}}

\newcommand{\Positive}{\ensuremath{\text{\it pos}}}
\newcommand{\At}{\ensuremath{\text{\it at}}}
\newcommand{\goleft}{\ensuremath{\text{\it goLeft}}}
\newcommand{\MaxInt}{\ensuremath{\text{\it maxInt}}}
\newcommand{\MinInt}{\ensuremath{\text{\it minInt}}}
\newcommand{\Underflow}{\ensuremath{\text{\it underflow}}}

\newcommand{\Happy}{\ensuremath{\text{\it happy}}}
\newcommand{\Sugar}{\ensuremath{\text{\it sugar}}}
\newcommand{\Salt}{\ensuremath{\text{\it salt}}}
\newcommand{\drink}{\ensuremath{\text{\it drink}}}

\newcommand{\married}{\ensuremath{\text{\it married}}}
\newcommand{\bachelor}{\ensuremath{\text{\it bachelor}}}

\newcommand{\Holds}[1]{\ensuremath{\text{\it Holds}(#1)}}

\newcommand{\Do}[1]{\ensuremath{\text{\it do}(#1)}}
\newcommand{\Poss}[1]{\ensuremath{\text{\it Poss}(#1)}}

\newcommand{\verum}{\ensuremath{\text{\it true}}}

\newcommand{\ActTheory}[2]{\tuple{\STAT{}{#2},\EFF{#1}{#2},\EXE{#1}{#2},
\INEXE{#1}{#2}}}

\newcommand{\STAT}[2]{\ensuremath{\mathcal{S}^{#1}_{#2}}}
\newcommand{\EFF}[2]{\ensuremath{\mathcal{E}^{#1}_{#2}}} 
\newcommand{\EFFp}[2]{\ensuremath{{\mathcal{E}}^{{#1}+}_{#2}}}
\newcommand{\EFFm}[2]{\ensuremath{{\mathcal{E}}^{{#1}-}_{#2}}}
\newcommand{\EXE}[2]{\ensuremath{\mathcal{X}^{#1}_{#2}}}
\newcommand{\INEXE}[2]{\ensuremath{\mathcal{I}^{#1}_{#2}}}
\newcommand{\IMPSTAT}[1]{\ensuremath{{\mathcal{S}_{\text{{\it imp}}}}_{#1}}}

\newcommand{\IMPINEXE}[2]
{\ensuremath{{{\mathcal{I}^{#1}_{{\text{{\it imp}}}}}_{#2}}}}
\newcommand{\STATnew}[1]{\STAT{}{\text{{\it new}}}}
\newcommand{\IMPSTATtotal}
{\ensuremath{\mathcal{S}_{{\text{{\it imp}*}}}}}
\newcommand{\CONSEQ}[2]{\ensuremath{\mathcal{C}^{#1}_{#2}}}
\newcommand{\FRA}[2]{\ensuremath{\mathcal{F}^{#1}_{#2}}}

\newcommand{\THE}[2]{\ensuremath{\mathcal{T}^{#1}_{#2}}}

\newcommand{\true}{1}

\newcommand{\wMvalid}[2]{\models_{\raisebox{-0.2ex}{$\!\!_{#1}$}}^
{\raisebox{-0.3ex}{$\!\!^{#2}$}}}
\newcommand{\notwMvalid}{\not\wMvalid}
\newcommand{\Mvalid}[1]{\models^{\raisebox{-0.3ex}{$\!\!^{#1}$}}}
\newcommand{\notMvalid}{\not\Mvalid}
\newcommand{\Cvalid}[1]{\models_{\raisebox{-0.25ex}{$\!\!_{#1}$}}}
\newcommand{\notCvalid}{\not\Cvalid}

\newcommand{\CPL}{{\sf CPL}}
\newcommand{\CPLvalid}{\Cvalid{\CPL}}

\newcommand{\K}{{\sf K}}

\newcommand{\Sfour}{{\sf S4}}

\newcommand{\PDL}{{\sf PDL}}
\newcommand{\PDLmodel}{\tuple{\Worlds,\AccRel,\Val}}
\newcommand{\PDLvalid}{\Cvalid{\PDL}}
\newcommand{\notPDLvalid}{\not\PDLvalid}

\newcommand{\EPDL}{{\sf EPDL}}
\newcommand{\EPDLvalid}{\Cvalid{\EPDL}}
\newcommand{\notEPDLvalid}{\not\EPDLvalid}

\newcommand{\depbinvalid}{\Cvalid{\depbin}}
\newcommand{\notdepbinvalid}{\not\depbinvalid}

\newcommand{\PI}[1]{{\it PI}(#1)} 
\newcommand{\NewCons}[2]{%
\ifthenelse{\equal{#1}{}\and\equal{#2}{}}{{\it NewCons}}
{{\it NewCons}_{#1}(#2)}
}

\newcommand{\depbin}{\leadsto}
\newcommand{\depbintuple}[1]{\tuple{#1}}
\newcommand{\indepbin}{\not\leadsto}

\newcommand{\causesif}[3]{\ensuremath{#1\ {\sf causes}\ #2\ {\sf if}\ #3}}

\newtheorem{definition}{Definition}[section]

\newtheorem{lemma}{Lemma}[section]
\newtheorem{theorem}{Theorem}[section]
\newtheorem{corollary}{Corolary}[section]
\newtheorem{example}{Example}[section]
\newtheorem{remark}{Remark}[section]

\newtheorem{algorithm}{Algorithm}[section]{\bfseries}{\rmfamily}
\makeatletter
\renewcommand{\@begintheorem}[2]{ 
        \trivlist\item[\hskip\labelsep{\bf #1\ #2}]}
\renewcommand{\@opargbegintheorem}[3]{\trivlist
        \item[\hskip \labelsep{\bf #1\ #2\ (#3)}]}
\makeatother
\newenvironment{proof}{\noindent {\bf Proof:}}{\hfill \qed}

\newcommand{\PostulConsist}{\ensuremath{\text{\bf PC}}}
\newcommand{\PostulStatLaw}{\ensuremath{\text{\bf PS}}}
\newcommand{\PostulInexeLaw}{\ensuremath{\text{\bf PI}}}
\newcommand{\PostulExeLaw}{\ensuremath{\text{\bf PX}}}
\newcommand{\PostulConsistGen}{\ensuremath{\text{\bf PC*}}}
\newcommand{\PostulStatLawGen}{\ensuremath{\text{\bf PS*}}}
\newcommand{\PostulInexeLawGen}{\ensuremath{\text{\bf PI*}}}

\newcommand{\PostulCompleteExe}{\ensuremath{\text{\bf PX$^+$}}}
\newcommand{\PostulEffLaw}{\ensuremath{\text{\bf PE}}}
\newcommand{\PostulAnatainEff}{\ensuremath{\text{\bf P$\bot$}}}

\newcommand{\A}{\ensuremath{\mathcal{A}}}
\newcommand{\AR}{\ensuremath{\mathcal{AR}}}

\newcommand{\myskip}{\medskip}

\newcounter{postulate}
\newcounter{proof_item}
\title{Metatheory of actions: beyond consistency}
\author{Andreas Herzig \hspace*{0.5cm} Ivan Varzinczak \\
IRIT -- 118  route de Narbonne \\ 31062 Toulouse Cedex -- France\\
e-mail: \url{{herzig,ivan}@irit.fr}\\
\url{http://www.irit.fr/recherches/LILAC}}

\begin{document}

\maketitle

\begin{abstract}
Consistency check has been the only criterion for theory evaluation in
logic-based approaches to reasoning about actions. This work goes beyond that
and contributes to the metatheory of actions by investigating what other
properties a good domain description in reasoning about actions should have. We 
state some metatheoretical postulates concerning this sore spot. When all
postulates are satisfied together we have a modular action theory. Besides being
easier to understand and more elaboration tolerant in McCarthy's sense, modular
theories have interesting properties. We point out the problems that arise when
the postulates about modularity are violated and propose algorithmic checks that
can help the designer of an action theory to overcome them.
\end{abstract}

\newpage

\tableofcontents

\newpage

\section{Introduction}\label{intro}

  In logic-based approaches to knowledge representation, a given domain is
described by a set of logical formulas $\THE{}{}$, which we call a (non-logical)
{\em theory}. That is also the case for reasoning about actions, where we are
interested in theories describing particular actions. We call such theories {\em
action theories}.

  A priori satisfiability is the only criterion that formal logic provides to
check the quality of such descriptions. In this work we go beyond that, and
argue that we should require more than the mere existence of a model for a given
theory.

  Our starting point is that in reasoning about actions one usually
distinguishes several kinds of logical formulas. Among these are effect axioms,
precondition axioms, and domain constraints. In order to distinguish such
non-logical axioms from logical axioms, we prefer to speak of effect laws,
executability laws, and static laws, respectively. Moreover we single out those
effect laws whose effect is $\bot$, and call them inexecutability laws.

  Given these types of laws, suppose the language is powerful enough to state
that action $\act$ is inexecutable in contexts where $\fml_{1}$ holds, and
executable in contexts where $\fml_{2}$ holds. It follows that there can be no
context where $\fml_{1}\et\fml_{2}$ holds. Now $\neg(\fml_{1}\et\fml_{2})$ is a
static law that does not mention $\act$. It is natural to expect that
$\neg(\fml_{1}\et\fml_{2})$ follows from the static laws alone. By means of
examples we show that when this is not the case, then unexpected conclusions
might follow from the theory $\THE{}{}$, even in the case $\THE{}{}$ is
consistent.

  This motivates postulates requiring that the different laws of an action
theory should be arranged modularly, \ie, in separated components, and in such a
way that interactions between them are limited and controlled. In essence, we
argue that static laws may influence the laws for actions, but the dynamic part
of a theory should not influence the non-dynamic one. It will turn out that in
all existing accounts allowing for these four kinds of
laws~\cite{Lin95,McCainTurner95,Thielscher95,CastilhoEtAl99,ZhangFoo01},
consistent action theories can be written that violate this requirement. We here
give algorithms that allow one to check whether an action theory satisfies the
postulates we state. With such algorithms, the task of correcting flawed action
theories can be made easier.

\myskip

  Although we here use the syntax of propositional dynamic logic
(\PDL)~\cite{Harel84}, all we shall say applies as well to first-order
formalisms, in particular to the Situation Calculus~\cite{McCarthyHayes69}. All
postulates we are going to present can be stated as well for other frameworks,
in particular for action languages such as $\A$,
$\AR$~\cite{GelfondLifschitz93,KarthaLifschitz94,GiunchigliaEtAl97} and others,
and for Situation Calculus based approaches. In~\cite{HerzigVarzinczak-IJCAI05}
we have given a Situation Calculus version of our analysis.

\myskip

  This work is organized as follows: after some background definitions
(Section~\ref{prelim}) we state (Section~\ref{postulates}) some postulates
concerning action descriptions. In Sections~\ref{no_impstat}
and~\ref{no_inexec}, we study the two most important of these postulates, giving
algorithmic methods to check whether an action theory satisfies them or not. We
then generalize (Section~\ref{generalizing}) and discuss
(Section~\ref{disturbing}) possible strengthenings of our set of postulates, and
show interesting results that their satisfaction gives us
(Section~\ref{discussion}). Finally, before concluding, we assess related work 
found in the literature on metatheory of actions (Section~\ref{rel_works}). 

\section{Preliminaries}\label{prelim}

\subsection{Dynamic logic}\label{dynamic_logic}

  Here we establish the ontology of dynamic domains. As our base formalism we
use \PDL. For more details, see~\cite{Harel84,HarelEtAl00}.

\myskip

  Let $\Act=\{\act_{1},\act_{2},\ldots\}$ be the set of all {\em atomic action
constants} of a given domain. Examples of atomic actions are $\load$ and
$\shoot$. We use $\act$ as a variable standing for a particular atomic
action. To each atomic action $\act$ there is an associated modal operator
$\nec\act$. Here we suppose that the underlying multimodal logic is
independently axiomatized (\ie, the logic is a fusion and there is no
interaction between the modal operators~\cite{KrachtWolter91,KrachtWolter97}).

  $\Prp=\{\prp_{1},\prp_{2},\ldots\}$ denotes the set of all {\em propositional
constants}, also called {\em fluents} or {\em atoms}. Examples of those are
$\Loaded$ and $\Alive$. We use $\prp$ as an atom variable.

  We suppose both $\Act$ and $\Prp$ are finite.

  We use small Greek letters $\fml,\psi,\ldots$ to denote {\em classical
formulas}. They are recursively defined in the following way:
\[
\fml\ ::= p\ |\ \top\ |\ \bot\ |\ \neg\fml\ |\ \fml\et\fml\ |\ \fml\ou\fml\ |\
\fml\imp\fml\ |\ \fml\sii\fml\
\]$\Fml$ is the set of all classical formulas.

  Examples of classical formulas are $\Walking\imp\Alive$ and
$\neg(\bachelor\et\married)$.

  A classical formula is {\em classically consistent} if there is at least one
valuation in the classical propositional logic that makes it true. Given
$\fml\in\Fml$, $\val(\fml)$ denotes the set of all valuations of $\fml$. We
identify $\models$ with the logical consequence in Classical Propositional
Logic~$\CPLvalid$.

  The set of all literals is $\Lit=\Prp\cup\{\neg\prp: \prp\in\Prp\}$. Examples
of literals are $\Alive$ and $\neg\Walking$. $\lit$ will be used as a literal
variable. If $\lit=\neg\prp$, then we identify $\neg\lit$ with $\prp$. 

  A {\em clause} $\clause$ is a disjunction of literals. We say that a literal
$\lit$ {\em appears} in a clause~$\clause$, written $\lit\in\clause$, if $\lit$
is a disjunct of $\clause$.

  We denote complex formulas (with modal operators) by capital Greek letters
$\modfml_{1},\modfml_{2},\ldots$ They are recursively defined in the following 
way:
\[
\modfml\ ::= \fml\ |\ \nec\act\modfml\ |\ \poss\act\modfml\ |\
\neg\modfml\ |\ \modfml\et\modfml\ |\ \modfml\ou\modfml\ |\ \modfml\imp\modfml\
|\ \modfml\sii\modfml\ 
\]where $\modfml$ denotes a complex formula. $\poss\act$ is the dual operator of
$\nec\act$, defined as
$\poss\act\modfml\defined\neg\nec\act\neg\modfml$. Sequential composition of
actions is defined by the abbreviation
$\nec{\act_{1};\act_{2}}\modfml\defined\nec{\act_{1}}\nec{\act_{2}}\modfml$.
Examples of complex formulas are $\Loaded\imp\nec\shoot\neg\Alive$ and
$\nec\load\Loaded$.

  For parsimony's sake, whenever there is no confusion we identify a set of
formulas with the conjunction of the formulas it is made of. The semantics we
take into account here is that for
multimodal~\K~\cite{Popkorn94,BlackburnEtAl01}.

\begin{definition}
A {\em \PDL-model} is a triple $\model=\PDLmodel$ where $\Worlds$ is a nonempty
set of possible worlds (alias possible states), $\AccRel$: $\Act\longrightarrow
2^{\Worlds\times\Worlds}$ maps action constants $\act$ to accessibility
relations $\ract\subseteq\Worlds\times\Worlds$, and $\Val$: $\Prp\longrightarrow
2^{\Worlds}$ maps propositional constants to subsets of $\Worlds$.
\end{definition}

\begin{definition}
\label{truth_conditions-PDL}
Given a \PDL-model $\model=\PDLmodel$, the satisfaction relation is defined as
the smallest relation satisfying:
\begin{itemize}
\item $\wMvalid{w}{\model}\prp$ ($\prp$ is true at world $w$ of model $\model$)
if $w\in\Val(\prp)$;
\item $\wMvalid{w}{\model}\nec\act\modfml$ if for every $w'$ such that $w\ract
w'$, $\wMvalid{w'}{\model}\modfml$;
\item the usual truth conditions for the other connectives.
\end{itemize}
\end{definition}

\begin{definition}
\label{PDL-model_of_formulas}
A \PDL-model $\model$ is a model of $\modfml$ (noted $\Mvalid{\model}\modfml$)
if and only if for all $w\in\Worlds$, $\wMvalid{w}{\model}\modfml$. $\model$ is
a model of a set of formulas $\THE{}{}$ (noted $\Mvalid{\model}\THE{}{}$) if and
only if $\wMvalid{w}{\model}\modfml$ for every $\modfml\in\THE{}{}$.
\end{definition}

\begin{definition}
A formula $\modfml$ is a {\em consequence of the set of global axioms}
$\{\modfml_{1},\ldots,\modfml_{n}\}$ in the class of all \PDL-models (noted
$\{\modfml_{1},\ldots,\modfml_{n}\}\PDLvalid\modfml$) if and only if for every
\PDL-model $\model$, if $\Mvalid{\model}\modfml_{i}$ for every $\modfml_{i}$,
then $\Mvalid{\model}\modfml$.\footnote{In~\cite{CastilhoEtAl99} local
consequence is considered. For that reason a further modal operator $\Box$ had
to be introduced, resulting in a logic which is multimodal \K\ plus monomodal
\Sfour\ for $\Box$, and where axiom schema $\Box\modfml\imp\nec\act\modfml$
holds.}
\end{definition}

  Having established the formal substratum our presentation will rely on, we
present in the next section the different types of formulas we use to describe
dynamic domains.

\subsection{Describing action theories in \PDL}\label{descrATinPDL}

  Before elaborating a theory, we need to specify what we are about to describe,
\ie, what the formulas we state talk about. Following the tradition in the
literature, we identify a domain (alias scenario) with the actions we take into
account and the fluents they can change. More formally, we have:

\begin{definition}
A {\em domain} is a tuple $\tuple{\Act,\Prp}$.
\end{definition}

  An example of a domain is the well-known Yale Shooting
Scenario~\cite{HanksMcDermott86}, whose actions are \load, \wait\ and \shoot,
and whose fluents are \Loaded\ and \Alive.

  Given a domain, we are interested in theories whose statements describe the
behavior of actions on the considered fluents. \PDL\ allows for the
representation of such statements, that we here call {\em action laws}. We
distinguish several types of them. We call {\em effect laws} formulas relating
an action to its effects. Statements of conditions under which an action cannot
be executed are called {\em inexecutability laws}. {\em Executability laws} in
turn stipulate the context where an action is guaranteed to be
executable. Finally, {\em static laws} are formulas that do not mention
actions. They express constraints that must hold in every possible state. These
four types of laws are our fundamental entities and we introduce them more
formally in the sequel.

\subsubsection{Static laws}

  Frameworks which allow for indirect effects of actions make use of logical
formulas that state invariant propositions about the world. Such formulas
delimit the set of possible states. They do not refer to actions, and we
suppose here that they are expressed as formulas of classical propositional
logic.

\begin{definition}
A {\em static law}\footnote{Static laws are often called {\em domain
constraints} or {\em integrity constraints}. Because the different laws for
actions that we shall introduce in the sequel could in principle also be called
like that, we avoid these terms.} is a formula $\fml\in\Fml$ that is classically
consistent.
\end{definition}

  An example of a static law is $\Walking\imp\Alive$, saying that if a turkey
is walking, then it must be alive~\cite{Thielscher95}. Another one is
$\Saved\sii(\Mbox1\ou\Mbox2)$, which states that an e-mail message is saved if
and only if it is in mailbox 1 or in mailbox 2 or both~\cite{CastilhoEtAl02}.

\myskip

  In action languages such as $\A$ and $\AR$ we would write the statement
$\Alive\text{ {\sf if} }\Walking$, and in the Situation Calculus it would be the
first-order formula
\[
\forall s (\Holds{\Walking,s}\imp\Holds{\Alive,s}).
\]

  The set of all static laws of a given domain is denoted by $\STAT{}{}$. At
first glance, no requirement concerning consistency of $\STAT{}{}$ is made. Of
course, we want $\STAT{}{}$ to be consistent, otherwise the whole theory is
inconsistent. As we are going to see in the sequel, however, consistency of
$\STAT{}{}$ alone is not enough to guarantee the consistency of a theory.

\subsubsection{Effect laws}

  Logical frameworks for reasoning about actions contain expressions linking
actions and their effects. We suppose that such effects might be conditional,
and thus get a third component of such laws.

  In \PDL, the formula $\nec\act\fml$ expresses that $\fml$ is true after every
possible execution of $\act$.

\begin{definition}
An {\em effect law\footnote{Effect laws are often called {\em action laws}, but
we prefer not to use that term here because it would also apply to executability
laws that are to be introduced in the sequel.} for action} $\act$ is of the form
$\antec\imp\nec\act\conseq$, where $\antec,\conseq\in\Fml$, with $\antec$ and
$\conseq$ both classically consistent.
\end{definition}
The consequent $\conseq$ is the effect which obtains when action $\act$ is
executed in a state where the antecedent $\antec$ holds. An example of an effect
law is $\Loaded\imp\nec\shoot\neg\Alive$, saying that whenever the gun is
loaded, after shooting the turkey is dead. Another one is
$\top\imp\nec\tease\Walking$: in every circumstance, the result of teasing is
that the turkey starts walking. For parsimony's sake, the latter effect law will
be written $\nec\tease\Walking$.

  Note that the consistency requirements for $\antec$ and $\conseq$ make sense:
if $\antec$ is inconsistent then the effect law is superfluous; if $\conseq$ is
inconsistent then we have an inexecutability law, that we consider as a separate
entity and which we are about to introduce formally in the sequel.

\myskip

  For the first example above, in action languages one would write the statement
\[
\causesif{shoot}{\neg\Alive}{\Loaded},
\]and in the Situation Calculus formalism one would write the first-order
formula
\[
\forall s (\Holds{\Loaded,s}\imp\neg\Holds{\Alive,\Do{\shoot, s}}).
\]

\subsubsection{Inexecutability laws}

  We consider effect laws whose consequent $\conseq$ is inconsistent as a
particular kind of law which we call inexecutability laws. (Such laws are
sometimes called qualifications~\cite{McCarthy77}.) This
allows us to avoid mixing things that are conceptually different: for an action
$\act$, an effect law mainly associates it with a consequent $\conseq$, while an
inexecutability law only associates it with an antecedent $\antec$, \viz\ the
context which precludes the execution of $\act$.

\begin{definition}
An {\em inexecutability law for action} $\act$ is of the form
$\antec\imp\nec\act\bot$, where $\antec\in\Fml$ is classically consistent.
\end{definition}

  For example $\neg\HasGun\imp\nec\shoot\bot$ expresses that $\shoot$ cannot be
executed if the agent has no gun. Another example is $\Dead\imp\nec\tease\bot$:
a dead turkey cannot be teased.

\myskip

  In $\AR$ we would write the statement $\text{{\sf impossible}
}\shoot\text{ {\sf if} }\neg\HasGun$, and in the Situation Calculus our example
would be
\[
\forall s (\neg\Holds{\HasGun,s}\imp\neg\Poss{\shoot, s}).
\]

\subsubsection{Executability laws}

  With only static and effect laws one cannot guarantee that the action $\shoot$
can be executed whenever the agent has a gun. We need thus a way to state the
conditions under which an action is guaranteed to be executable.

  In dynamic logic the dual $\poss\act\fml$, defined as $\neg\nec\act\neg\fml$,
can be used to express executability. $\poss\act\top$ thus reads ``the execution
of action $\act$ is possible''.

\begin{definition}
An {\em executability law\footnote{Some approaches (most prominently Reiter's)
use biconditionals $\fml\sii\poss\act\top$, called precondition axioms. This is
equivalent to $\neg\fml\sii\nec\act\bot$, highlighting that they merge
information about inexecutability with information about executability. In this
work we consider these entities different and keep them separated.} for action}
$\act$ is of the form $\antec\imp\poss\act\top$, where $\antec\in\Fml$ is
classically consistent.
\end{definition}

  For instance $\HasGun\imp\poss\shoot\top$ says that shooting can be executed
whenever the agent has a gun, and $\top\imp\poss\tease\top$, also written
$\poss\tease\top$, establishes that the turkey can always be teased.

\myskip

  In action languages such laws are not represented. In Situation Calculus our
example would be stated as
\[
\forall s (\Holds{\HasGun,s}\imp\Poss{\shoot, s}).
\]

  Whereas all the extant approaches in the literature that allow for indirect
effects of actions contain static and effect laws, and provide a way for
representing inexecutabilities (in the form of implicit
qualifications~\cite{GinsbergSmith88,Lin95,Thielscher95}), the status of
executability laws is less consensual. Some
authors~\cite{Schubert90,DohertyEtAl96,McCainTurner95,Thielscher95} more or less
tacitly consider that executability laws should not be made explicit but rather
inferred by the reasoning mechanism. Others~\cite{Lin95,ZhangFoo01} have
executability laws as first class objects one can reason about.

  It seems a matter of debate whether one can always do without
executabilities. In principle it seems to be strange to just state information
about necessary conditions for action execution (inexecutabilities) without
saying anything about its sufficient conditions. This is the reason why we think
that we need executability laws. Indeed, in several domains one wants to
explicitly state under which conditions a given action is guaranteed to be
executable, \eg\ that a robot never gets stuck and is always able to execute a
move action. And if we have a plan such as $\load;\shoot$ (\load\ followed by
\shoot) of which we know that it achieves the goal $\neg\Alive$, then we would
like to be sure that it is executable in the first place!\footnote{Of course
this would require a solution to the qualification problem~\cite{McCarthy77}.}
In any case, allowing for executability laws gives us more flexibility and
expressive power.

\subsubsection{Action theories}

  Given a domain $\tuple{\Act,\Prp}$, for an action~$\act\in\Act$, we define
$\EFF{\act}{}$ as the set of its effect laws, $\EXE{\act}{}$ the set of its
executability laws, and $\INEXE{\act}{}$ that of its inexecutability laws.

\begin{definition}
An {\em action theory for $\act$} is a tuple $\THE{\act}{}=\ActTheory{\act}{}$.
\end{definition}

  In our running scenario example, a theory for the action \shoot\ would be
\[
\STAT{}{}=\{ \Walking\imp\Alive \},\text{ }
\EFF{\shoot}{}=\{ \Loaded\imp\nec\shoot\neg\Alive \},
\]\[
\EXE{\shoot}{}=\{ \HasGun\imp\poss\shoot\top \},\text{ }
\INEXE{\shoot}{}=\{ \neg\HasGun\imp\nec\shoot\bot \}
\]

  Given a dynamic domain we define $\EFF{}{}=\bigcup_{\act\in\Act}\EFF{\act}{}$,
$\EXE{}{}=\bigcup_{\act\in\Act}\EXE{\act}{}$, and $\INEXE{}{}=\bigcup_{\act\in
\Act}\INEXE{\act}{}$. All these sets are finite, because $\Act$ is finite and
each of the $\EFF{\act}{}$, $\EXE{\act}{}$, $\INEXE{\act}{}$ is finite.

\begin{definition}
An {\em action theory} $\THE{}{}$ is a tuple of the form $\ActTheory{}{}$.
\end{definition}

  For parsimony's sake, whenever there is no confusion we write
$\STAT{}{},\EFF{}{},\EXE{}{},\INEXE{}{}\PDLvalid\modfml$ instead of
$\STAT{}{}\cup\EFF{}{}\cup\EXE{}{}\cup\INEXE{}{}\PDLvalid\modfml$.

  When performing the task of formalizing dynamic domains, we face the {\em
frame problem}~\cite{McCarthyHayes69} and the {\em ramification
problem}~\cite{Finger87}. In what follows we formally present the logic of
actions in which action theories will henceforth be described.

\subsection{Dynamic logic and the frame problem}\label{DynamicLogic}

  As it was already expected, the logical formalism of \PDL\ alone does not
solve the frame problem. For instance, if $\ActTheory{}{}$ describes our
shooting domain, then
\[
\STAT{}{},\EFF{}{},\EXE{}{},\INEXE{}{}\notPDLvalid\HasGun\imp\nec\load\HasGun.
\]

  The same can be said about the ramification problem in what concerns the
derivation of indirect effects not properly caused by the action under
consideration. For example,
\[
\STAT{}{},\EFF{}{},\EXE{}{},\INEXE{}{}\PDLvalid\nec\tease\Alive.
\]

  Thus, given an action theory $\ActTheory{}{}$, we need a consequence relation
powerful enough to deal with the frame and ramification problems. This means
that the deductive power of \PDL\ has to be augmented in order to ensure that
the only non-effects of actions that follow from the theory are those that are
really relevant. The presence of static constraints makes that this is a
delicate task, and starting with~\cite{Lin95,McCainTurner95}, several authors
have argued that some notion of causality is needed. We here opt for the
dependence based approach presented in~\cite{CastilhoEtAl99}, which has been 
shown in~\cite{DemolombeEtAl-JANCL03} to subsume Reiter's solution to the frame
problem~\cite{Reiter91}, and moreover at least partially accounts for the
ramification problem.

  In the logical framework developed in~\cite{CastilhoEtAl99}, metalogical
information, given in the form of a dependence relation, is added to~\PDL.

\begin{definition}[Dependence relation~\cite{CastilhoEtAl99}]
A {\em dependence relation} is a binary relation
$\depbin\ \subseteq\Act\times\Lit$.
\end{definition}

  The expression $\act\depbin\lit$ denotes that the execution of action $\act$
{\em may make} the literal $\lit$ true. In our example we have
\[
\depbin\ = \left\{
    \begin{array}{c}
       \depbintuple{\shoot,\neg\Loaded},
       \depbintuple{\shoot,\neg\Alive},\\ 
       \depbintuple{\shoot,\neg\Walking}, 
       \depbintuple{\tease,\Walking}
    \end{array}
    \right\},
\]
which means that action \shoot\ may make the literals $\neg\Loaded$,
$\neg\Alive$ and $\neg\Walking$ true, and action \tease\ may make \Walking\
true.

  Semantically, the dependence-based approach relies on the explanation closure
assumption~\cite{Schubert90}. The reasoning behind its solution to the frame
problem consists in a kind of negation as failure: Because
$\depbintuple{{\load},\neg{\HasGun}}\notin\ \depbin$, we have 
$\load\indepbin\neg\HasGun$, \ie, $\neg\HasGun$ is never caused by \load. Thus,
in a context where $\HasGun$ is true, after every execution of $\load$,
$\HasGun$ still remains true. We also have $\tease\indepbin\Alive$ and
$\tease\indepbin\neg\Alive$. The meaning of all these independences is that the
frame axioms $\HasGun\imp\nec\load\HasGun$, $\neg\Alive\imp\nec\tease\neg\Alive$
and $\Alive\imp\nec\tease\Alive$ hold.

  We assume $\depbin$ is finite.

\myskip

  A dependence relation $\depbin$ defines a class of possible worlds models
$\modelclass{\depbin}$.

\begin{definition}
\label{truth_conditions-depbin}
Given a $\depbin$-model $\model=\PDLmodel$, the satisfaction relation is defined
as the smallest relation satisfying:
\begin{itemize}
\item all the truth conditions of Definition~\ref{truth_conditions-PDL}.
\item whenever $w\ract w'$ then:
\begin{itemize}
\item $\notwMvalid{w}{\model}\prp$ implies $\notwMvalid{w'}{\model}\prp$, if
$\act\indepbin\prp$;
\item $\wMvalid{w}{\model}\prp$ implies $\wMvalid{w'}{\model}\prp$, if
$\act\indepbin\neg\prp$.
\end{itemize}
\end{itemize}
\end{definition}

  Given $\model\in\modelclass{\depbin}$, $\modfml$ and $\THE{}{}$,
$\Mvalid{\model}\modfml$ and $\Mvalid{\model}\THE{}{}$ are defined as in
Definition~\ref{PDL-model_of_formulas}.

\begin{definition}
A formula $\modfml$ is a {\em $\depbin$-based consequence} of
$\{\modfml_{1},\ldots,\modfml_{n}\}$ in the class of all $\depbin$-models (noted
$\{\modfml_{1},\ldots,\modfml_{n}\}\depbinvalid\modfml$) if and only if for
every $\depbin$-model $\model$, if $\Mvalid{\model}\modfml_{i}$ for every
$\modfml_{i}$, then $\Mvalid{\model}\modfml$.
\end{definition}

  In our example it thus holds
\[
\STAT{}{},\EFF{}{},\EXE{}{},\INEXE{}{}\depbinvalid\HasGun\imp\nec\load\HasGun
\]and
\[
\STAT{}{},\EFF{}{},\EXE{}{},\INEXE{}{}\depbinvalid\neg\Alive\imp\nec\tease\neg
\Alive.
\]

  In this way, the dependence-based approach solves the frame problem. However,
it does not entirely solve the ramification problem: while indirect effects such
as $\Loaded\imp\nec\shoot\neg\Walking$ can be deduced with $\depbinvalid$
without explicitly stating that {\em in the set of effect laws} for \shoot, we
still have to state {\em indirect dependences} such as
$\shoot\depbin\neg\Walking$. Nevertheless, according to Reiter's view:

\begin{quote}
``what counts as a solution to the frame problem $\ldots$ is a systematic
procedure for generating, from the effect laws, $\ldots$ a parsimonious
representation for [all] the frame axioms''~\cite{Reiter01}.
\end{quote}

  We comply with that as we can define a semi-automatic procedure for generating
the dependence relation from the set of effect laws. Moreover, as it has been
argued in~\cite{CastilhoEtAl02,HerzigVarzinczak-TechRep04}, our approach is in
line with the state of the art because none of the existing solutions to the
frame and the ramification problems can handle domains with both indeterminate
and indirect effects.

\myskip

  In the next section we turn to a metatheoretical analysis of action theories
and make a step toward formal criteria for theory evaluation. Before that, we
need a definition.

\begin{definition}
Let $\ActTheory{}{}$ be an action theory and $\depbin$ its associated dependence
relation. Then $\model=\PDLmodel$ is the {\em big} (alias {\em maximal/standard})
{\em model} for $\ActTheory{}{}$ and $\depbin$ if and only if:
\begin{itemize}
\item $\model$ is a $\depbin$-model;
\item $\Worlds=\val(\STAT{}{})$ (all valuations of \STAT{}{});
\item $\ract=\{(w,w'):\forall\antec\imp\nec\act\conseq\in\EFF{}{}\cup
\INEXE{}{},\text{ if } \wMvalid{w}{\model}\antec,\text{ then }
\wMvalid{w'}{\model}\conseq\}$.
\end{itemize}
\end{definition}

  In the rest of the paper we characterize when an action theory with a
dependence relation has a big model.

\section{Postulates}\label{postulates}

  ``When does a given action theory have a model?'', and, more importantly, ``is
that model what we really expect from it?'' are questions that naturally arise
when we talk about action theories. Here we claim that all the approaches that
are put forward in the literature are too liberal in the sense that we can have
satisfiable action theories that are intuitively incorrect. We argue that
something beyond the consistency notion is required in order to help us in
answering those questions.

  We do not attempt here to provide a `magical' method for making an action
theory intuitive. Instead, what we are going to do in what follows is to provide
some guidelines that help detecting unintuitive consequences of a theory and
identifying its problematic part(s). 

  Our central thesis is that the different types of laws define in
Section~\ref{descrATinPDL} should be neatly separated in different
modules. Besides that, we want such laws to interfere only in one sense: static
laws together with action laws for $\act$ may have consequences that do not
follow from the action laws for $\act$ alone. The other way round, action laws
should not allow to infer new static laws, effect laws should not allow to infer
inexecutability laws, action laws for $\act$ should not allow to infer action
laws for $\act'$, etc. This means that our logical modules should be designed in
such a way that they are as specialized and as little dependent on others as
possible.

  A first step in this direction has been the proposed division of our entities
into the sets $\STAT{}{}$, $\EFF{}{}$, $\EXE{}{}$ and $\INEXE{}{}$. In order to
accomplish our goal, we have to diminish interaction among such modules,
rendering them the least interwoven we can. The rest of the section contains
postulates expressing this.

\begin{itemize}
\item[] \PostulConsist\ ({\bf Logical consistency}):
$\STAT{}{},\EFF{\act}{},\EXE{\act}{},\INEXE{\act}{}\notdepbinvalid\bot$
\end{itemize}
The theory of a given action should be logically consistent.

\begin{itemize}
\item[] \PostulStatLaw\ ({\bf No implicit static laws}):
\[
\mbox{if }\ 
\STAT{}{},\EFF{\act}{},\EXE{\act}{},\INEXE{\act}{}\depbinvalid\fml, 
\mbox{ then }\ \STAT{}{}\models\fml
\]
\end{itemize}
If a classical formula can be inferred from the action theory, then it should be
inferable from the set of static laws alone. (Note that on the left we use
consequence in $\modelclass{\depbin}$, while on the right we use consequence in
classical logic: as both $\STAT{}{}$ and $\fml$ are classical, $\fml$ should be
inferable from $\STAT{}{}$ in classical logic.)

\begin{itemize}
\item[] \PostulInexeLaw\ ({\bf No implicit inexecutability laws}):
\[
\mbox{if }\
\STAT{}{},\EFF{\act}{},\EXE{\act}{},\INEXE{\act}{}\depbinvalid
\antec\imp\nec\act\bot, \mbox{ then }\
\STAT{}{},\INEXE{\act}{}\PDLvalid\antec\imp\nec\act\bot
\]
\end{itemize}
If an inexecutability law for an action $\act$ can be inferred from its action
theory, then it should be inferable in \PDL\ from the static laws and the
inexecutability laws for $\act$ alone. Note that we used $\PDLvalid$ instead of
$\depbinvalid$ because we also suppose that neither frame axioms nor indirect
effects should be relevant to derive inexecutability laws. The same remark holds
for the postulates that follow.

\begin{itemize}
\item[] \PostulExeLaw\ ({\bf No implicit executability laws}):
\[
\mbox{if }\
\STAT{}{},\EFF{\act}{},\EXE{\act}{},\INEXE{\act}{}\depbinvalid\antec\imp
\poss\act\top, \mbox{ then }\ \STAT{}{},\EXE{\act}{}\PDLvalid\antec\imp
\poss\act\top
\]
\end{itemize}
If an executability law for $\act$ can be inferred from its action theory, then
it should already ``be'' in~$\EXE{\act}{}$, in the sense that it should also be
inferable in \PDL\ from the set of static and executability laws for $\act$
alone.

\myskip

Postulate~\PostulConsist\ is obvious, for we are interested in consistent
theories. It can be shown that~\PostulExeLaw\ is a consequence of
\PostulStatLaw\ (see Corollary~\ref{PCconseqPS}).

\myskip

  Thus, while~\PostulConsist\ is obvious and~\PostulExeLaw\ can be ensured by
\PostulStatLaw, things are less obvious for Postulates~\PostulStatLaw\
and~\PostulInexeLaw: it turns out that for all approaches in the literature they
are easily violated by action theories that allow to express the four kinds of
laws. We therefore study each of these postulates in the subsequent sections by
means of examples, give algorithms to decide whether they are satisfied, and
discuss about what to do in the case the answer is `no'.

\section{No implicit static laws}\label{no_impstat}

  While executability laws increases expressive power, they might conflict with
inexecutability laws. Consider, for example, the following action theory:
\[
\STAT{}{1}=\{\Walking\imp\Alive\},\text{ }
\EFF{}{1}=\left\{
    \begin{array}{c}
          \nec\tease\Walking, \\
          \Loaded\imp\nec\shoot\neg\Alive
       \end{array}
     \right\},
\]\[
\EXE{}{1}=\{\poss\tease\top\},\text{ }
\INEXE{}{1}=\{\neg\Alive\imp\nec\tease\bot\}
\]and the dependence relation:
\[
\depbin\ = \left\{
    \begin{array}{c}
       \depbintuple{\shoot,\neg\Loaded},
       \depbintuple{\shoot,\neg\Alive},\\ 
       \depbintuple{\shoot,\neg\Walking}, 
       \depbintuple{\tease,\Walking}
    \end{array}
    \right\}
\]
From this description we have the unintuitive
$\EXE{\tease}{1},\INEXE{\tease}{1}\PDLvalid\Alive$: the turkey is immortal! This
is an {\em implicit static law} because $\Alive$ does not follow from
$\STAT{}{1}$ alone: $\ActTheory{\tease}{1}$ violates Postulate~\PostulStatLaw.

\myskip

  How can we find out whether an action theory for $\act$ satisfies
Postulate~\PostulStatLaw? 

\begin{theorem}\label{PS-BigModel}
$\ActTheory{}{}$ and $\depbin$ satisfy Postulate~\PostulStatLaw\ if and only if
the big model for $\ActTheory{}{}$ and $\depbin$ is a model of $\ActTheory{}{}$
and $\depbin$.
\end{theorem}
\begin{proof}

\noindent($\Rightarrow$): Let $\model=\PDLmodel$ be a big model of 
$\ActTheory{}{}$ and $\depbin$, and suppose $\Mvalid{\model}\STAT{}{}\et\EFF{}{}
\et\EXE{}{}\et\INEXE{}{}$ ($\model$ is a model of $\STAT{}{}\cup\EFF{}{}\cup
\EXE{}{}\cup\INEXE{}{}$). Then $\Worlds=\val(\STAT{}{})$, \ie, for all
$\fml\in\Fml$ and all $w\in\Worlds$, if $\wMvalid{w}{\model}\fml$, then there is
a valuation $v$ of $\STAT{}{}$ such that $v$ makes $\fml$ true. From this it
follows that if $\wMvalid{w}{\model}\fml$ for all $w\in\Worlds$, then $\fml$ is
true in all valuations of $\STAT{}{}$. Hence $\STAT{}{},\EFF{}{},\EXE{}{},
\INEXE{}{}\depbinvalid\fml$ implies $\STAT{}{}\models\fml$, and then
$\ActTheory{}{}$ and $\depbin$ satisfy Postulate~\PostulStatLaw.

\noindent($\Leftarrow$): Let $\model=\PDLmodel$ be a big model of
$\ActTheory{}{}$ and $\depbin$. Suppose $\ActTheory{}{}$ and $\depbin$ do not
satisfy Postulate~\PostulStatLaw. Then there must be $\fml\in\Fml$ such that 
$\STAT{}{},\EFF{}{},\EXE{}{},\INEXE{}{}\depbinvalid\fml$ and $\STAT{}{}\not
\models\fml$. This means that there is a valuation $v$ of $\STAT{}{}$ that
falsifies $\fml$. As $v\in\Worlds$ (because $\model$ is a big model) then
$\model$ is not a model of $\ActTheory{}{}$ and $\depbin$.
\end{proof}

  We shall give an algorithm to find a finite characterization of
all\footnote{Actually what the algorithm does is to find an interpolant of all
implicit static laws of the theory.} implicit static laws of a given action
theory $\ActTheory{\act}{}$. The idea is as follows: for each executability law
$\antec\imp\poss\act\top$ in the theory, construct from $\EFF{\act}{}$,
$\INEXE{\act}{}$ and $\depbin$ a set of inexecutabilities
$\{\antec_{1}\imp\nec\act\bot,\ldots,\antec_{n}\imp\nec\act\bot\}$ that 
potentially conflict with $\antec\imp\poss\act\top$. For each~$i$, $1\leq i\leq
n$, if $\antec\et\antec_{i}$ is satisfiable \wrt\ $\STAT{}{}$, mark
$\neg(\antec\et\antec_{i})$ as an implicit static law. Incrementally repeat this
procedure (adding all the $\neg(\antec\et\antec_{i})$ that were caught to
$\STAT{}{}$) until no implicit static law is obtained. 

  For an example of the execution of the algorithm, consider
$\ActTheory{\tease}{1}$ with $\depbin$ as above. For the action $\tease$, we
have the executability $\poss\tease\top$. Now, from $\EFF{\tease}{1}$,
$\INEXE{\tease}{1}$ and $\depbin$ we try to build an inexecutability for
$\tease$. We take $\nec\tease\Walking$ and compute then all indirect effects of
$\tease$ \wrt\ $\STAT{}{1}$. From $\Walking\imp\Alive$, we get that $\Alive$ is
an indirect effect of $\tease$, giving us $\nec\tease\Alive$. But 
$\depbintuple{\tease,\Alive}\notin\ \depbin$, which means the frame axiom
$\neg\Alive\imp\nec\tease\neg\Alive$ holds. Together with $\nec\tease\Alive$,
this gives us the inexecutability $\neg\Alive\imp\nec\tease\bot$. As
$\STAT{}{1}\cup\{\top,\neg\Alive\}$ is satisfiable ($\top$ is the antecedent of
the executability $\poss\tease\top$), we get $\neg\Alive\imp\bot$, \ie, the
implicit static law $\Alive$. For this example no other inexecutability for
$\tease$ can be derived, so the computation stops.

  Before presenting the pseudo-code of the algorithm we need some definitions.

\begin{definition}
Let $\fml\in\Fml$ and $\clause$ a clause. $\clause$ is an {\em implicate of}
$\fml$ if and only if $\fml\models\clause$.
\end{definition}

  In our running example, $\Alive$ is an implicate of the set of formulas
$\{\Walking\imp\Alive,\Walking\}$.

\begin{definition}
Let $\fml\in\Fml$ and $\clause$ a clause. $\clause$ is a {\em prime implicate of}
$\fml$ if and only if
\begin{itemize}
\item $\clause$ is an implicate of $\fml$, and
\item for every implicate $\clause'$ of $\fml$, $\clause'\models\clause$ implies
$\clause\models\clause'$.
\end{itemize}
The set of all prime implicates of a formula $\fml$ is denoted $\PI{\fml}$.
\end{definition}

  For example, the set of prime implicates of $\prp_{1}$ is just $\{\prp_{1}\}$,
and that of $\prp_{1}\et(\neg\prp_{1}\ou\prp_{2})\et(\neg\prp_{1}\ou\prp_{3}\ou
\prp_{4})$ is $\{\prp_{1}, \prp_{2}, \prp_{3}\ou\prp_{4}\}$. In our shooting
domain, $\Alive$ is a prime implicate of $\{\Walking\imp\Alive,\Walking\}$. For
more on prime implicates and their properties, see~\cite{Marquis2000}.

\begin{definition}
Let $\fml,\psi\in\Fml$. Then $\NewCons{\fml}{\psi}=\PI{\fml\et\psi}
\setminus\PI{\fml}$.
\end{definition}

  The function $\NewCons{\fml}{\psi}$ computes the {\em new consequences} of
$\fml$ \wrt\ $\psi$: the set of strongest clauses that follow from
$\fml\et\psi$, but do not follow from $\fml$ alone (\cf\ \eg~\cite{Inoue92}). It
is computed by subtracting the prime implicates of $\fml$ from those of
$\fml\et\psi$. For example, $\NewCons{\prp_{1}}{(\neg\prp_{1}\ou\prp_{2})\et
(\neg\prp_{1}\ou\prp_{3}\ou\prp_{4})}=\{\prp_{2},\prp_{3}\ou\prp_{4}\}$. And for
our scenario, $\NewCons{\Walking\imp\Alive}{\Walking}=\{\Alive,\Walking\}$.

\comment{
Given classical formulas $\fml_{1}$ and $\fml_{2}$, the function
$\NewCons{\fml_{1}}{\fml_{2}}$ computes the {\em new consequences} of $\fml_{2}$
\wrt\ $\fml_{1}$: the set of strongest clauses that follow from
$\fml_{1}\et\fml_{2}$, but do not follow from $\fml_{1}$ alone (\cf\
\eg~\cite{Inoue92}). It is known that $\NewCons{\fml_{1}}{\fml_{2}}$ can be
computed by subtracting the prime implicates of $\fml_{1}$ from those of
$\fml_{1}\et\fml_{2}$. For example, the set of prime implicates of $\prp_{1}$ is
just $\{\prp_{1}\}$, that of
$\prp_{1}\et(\neg\prp_{1}\ou\prp_{2})\et(\neg\prp_{1}\ou\prp_{3}\ou\prp_{4})$ is
$\{\prp_{1}, \prp_{2}, \prp_{3}\ou\prp_{4}\}$, hence
$\NewCons{\prp_{1}}{(\neg\prp_{1}\ou\prp_{2})\et(\neg\prp_{1}\ou\prp_{3}\ou
\prp_{4})}=\{\prp_{2},\prp_{3}\ou\prp_{4}\}$. And for our example,
$\NewCons{\Walking\imp\Alive}{\Walking}=\{\Alive,\Walking\}$.
}

  The algorithm below improves the one in~\cite{HerzigVarzinczak-Aiml04Proc05}
by integrating a solution to the frame problem (via the dependence relation
$\depbin$). As a matter of notation, we define 
$\CONSEQ{\act}{}=\EFF{\act}{}\cup\INEXE{\act}{}$ as the set of all formulas
expressing the direct consequences of an action $\act$, whether they are
consistent or not.

\begin{algorithm}[Finding all implicit static laws induced by $\act$]\mbox{
}\\ \label{alg_imp_stat}
\vspace*{-0.35cm}
\begin{algorithmic}
\INPUT $\ActTheory{\act}{}$ and $\depbin$
\OUTPUT $\IMPSTATtotal$, the set of all implicit static laws of
$\ActTheory{\act}{}$
\STATE $\IMPSTATtotal\GETS\ \emptyset$
\STATE $\CONSEQ{\act}{}\GETS\ \EFF{\act}{}\cup\INEXE{\act}{}$
\REPEAT
   \STATE $\IMPSTAT{}\GETS\ \emptyset$
      \FORALL{$\antec\imp\poss\act\top\in\EXE{\act}{}$}
         \FORALL{$\hat{\CONSEQ{\act}{}}\subseteq\CONSEQ{\act}{}$ such that
                 $\hat{\CONSEQ{\act}{}}\neq\emptyset$}
            \STATE $\antec_{\hat{\CONSEQ{\act}{}}}\GETS\ \bigwedge\{\antec_{i} :
                  \antec_{i}\imp\nec\act\conseq_{i} \in \hat{\CONSEQ{\act}{}}\}$
            \STATE $\conseq_{\hat{\CONSEQ{\act}{}}}\GETS\ \bigwedge\{\conseq_{i}:
                  \antec_{i}\imp\nec\act\conseq_{i} \in \hat{\CONSEQ{\act}{}}\}$
            \FORALL{$\clause\in\NewCons{\STAT{}{}}
                    {\conseq_{\hat{\CONSEQ{\act}{}}}}$}
               \IF{$\STAT{}{}\cup\IMPSTATtotal\cup
                   \{\antec, \antec_{\hat{\CONSEQ{\act}{}}},\neg\clause\}
                   \nvdash\bot$ \ALGAND\ 
                   $\forall\lit_{i}\in\clause ,\ \act\indepbin\lit_{i}$}
                  \STATE $\IMPSTAT{}\GETS\ \IMPSTAT{}\cup
                         \{\neg(\antec\et\antec_{\hat{\CONSEQ{\act}{}}}
                           \et\neg\clause)\}$
               \ENDIF
            \ENDFOR
         \ENDFOR
      \ENDFOR
   \STATE $\IMPSTATtotal\GETS\ \IMPSTATtotal\cup\IMPSTAT{}$
\UNTIL{$\IMPSTAT{}=\emptyset$}
\end{algorithmic}
\end{algorithm}

  In each step of the algorithm, $\STAT{}{}\cup\IMPSTATtotal$ is the updated set
of static laws (the original ones fed with the implicit laws caught up to that
point). At the end, $\IMPSTATtotal$ collects all the implicit static laws.

\begin{theorem}~\label{th_imp_stat_termination}
Algorithm~\ref{alg_imp_stat} terminates.
\end{theorem}
\begin{proof}
Let $\CONSEQ{\act}{}=\EFF{\act}{}\cup\INEXE{\act}{}$. First, the set of
candidates to be an implicit static law that might be due to $\act$ and that are
examined in the {\bf repeat}-loop is
\[
\{\neg(\antec\et\antec_{\hat{\CONSEQ{\act}{}}}\et\neg\clause):
\hat{\CONSEQ{\act}{}}\subseteq\CONSEQ{\act}{},\antec\imp\poss\act\top\in
\EXE{\act}{}\text{ and }\clause\in\NewCons{\STAT{}{}}
{\conseq_{\hat{\CONSEQ{\act}{}}}}\}
\]As $\EXE{\act}{}$ and $\INEXE{\act}{}$ are finite, this set is finite.

In each step either the algorithm stops because $\IMPSTAT{}=\emptyset$, or at
least one of the candidates is put into $\IMPSTAT{}$ in the outermost {\bf
for}-loop. (This one terminates, because $\EXE{\act}{}$, $\CONSEQ{\act}{}$ and
\NewCons{}{} are finite.) Such a candidate is not going to be put into
$\IMPSTAT{}$ in future steps, because once added to
$\STAT{}{}\cup\IMPSTATtotal$, it will be in the set of laws $\STAT{}{}\cup
\IMPSTATtotal$ of all subsequent executions of the outermost {\bf for}-loop,
falsifying its respective {\bf if}-test for such a candidate. Hence the {\bf
repeat}-loop is bounded by the number of candidates, and therefore
Algorithm~\ref{alg_imp_stat} terminates.
\end{proof}

\myskip

  This is the key algorithm of the paper. We are aware that it comes with
considerable computational costs: first, the number of formulas
$\antec_{\hat{\CONSEQ{\act}{}}}$ and $\conseq_{\hat{\CONSEQ{\act}{}}}$ is
exponential in the size of $\CONSEQ{\act}{}$, and second, the computation of
$\NewCons{\STAT{}{}}{\conseq_{\hat{\CONSEQ{\act}{}}}}$ might result in
exponential growth. While we might expect $\CONSEQ{\act}{}$ to be reasonably
small in practice (because $\EFF{\act}{}$ and $\INEXE{\act}{}$ are in general
small), the size of $\NewCons{\STAT{}{}}{\conseq_{\hat{\CONSEQ{\act}{}}}}$ is
more difficult to control.

\begin{example}\label{ex_imp_stat2}
For $\ActTheory{\tease}{1}$, Algorithm~\ref{alg_imp_stat} returns
$\IMPSTATtotal=\{\Alive\}$.
\end{example}

\begin{theorem}\label{th_imp_stat}
An action theory $\ActTheory{\act}{}$ with $\depbin$ satisfies
Postulate~\PostulStatLaw\ if and only if $\IMPSTATtotal=\emptyset$.
\end{theorem}
\begin{proof}
See Appendix~\ref{proof_th_imp_stat}.
\end{proof}

\begin{theorem}\label{th_imp_stat2}
Let $\IMPSTATtotal$ be the output of Algorithm~\ref{alg_imp_stat} on input
$\ActTheory{\act}{}$ and $\depbin$. Then
\begin{enumerate}
\item
$\tuple{\STAT{}{}\cup\IMPSTATtotal,\EFF{\act}{},\EXE{\act}{},\INEXE{\act}{}}$
has no implicit static law.\label{th_imp_stat2-it1}
\item $\STAT{}{},\EFF{\act}{},\EXE{\act}{},\INEXE{\act}{}\depbinvalid
\bigwedge\IMPSTATtotal$.\label{th_imp_stat2-it2}
\end{enumerate}
\end{theorem}
\begin{proof}
Item~\ref{th_imp_stat2-it1}.\ is straightforward from the termination of
Algorithm~\ref{alg_imp_stat} and
Theorem~\ref{th_imp_stat}. Item~\ref{th_imp_stat2-it2}.\ follows from the fact
that by the {\bf if}-test in Algorithm~\ref{alg_imp_stat}, the only formulas
that are put in $\IMPSTATtotal$ at each execution of the {\bf repeat}-loop are
exactly those that are implicit static laws of the original theory.
\end{proof}

\begin{corollary}~\label{theorem_imp_stat2}
For all $\fml\in\Fml$, $\STAT{}{},\EFF{\act}{},\EXE{\act}{},\INEXE{\act}{}
\depbinvalid\fml$ if and only if $\STAT{}{}\cup\IMPSTATtotal\models\fml$.
\end{corollary}
\begin{proof}
For the left-to-right direction, let $\fml\in\Fml$ be such that
$\STAT{}{},\EFF{\act}{},\EXE{\act}{},\INEXE{\act}{}\depbinvalid\fml$. Then
$\STAT{}{}\cup\IMPSTATtotal,\EFF{\act}{},\EXE{\act}{},\INEXE{\act}{}\depbinvalid
\fml$, by monotonicity. By Theorem~\ref{th_imp_stat2}-\ref{th_imp_stat2-it1}.,
$\tuple{\STAT{}{}\cup\IMPSTATtotal,\EFF{\act}{},\EXE{\act}{},\INEXE{\act}{}}$
has no implicit static law, hence $\STAT{}{}\cup\IMPSTATtotal\models\fml$.

\noindent The right-to-left direction is straightforward by
Theorem~\ref{th_imp_stat2}-\ref{th_imp_stat2-it2}.
\end{proof}

\myskip

  What shall we do once we have discovered an implicit static law?

  The existence of implicit static laws may indicate too strong executability
laws: in Example~\ref{ex_imp_stat2}, we wrongly assumed that \tease\ is always  
executable. Thus one way of `repairing' our theory would be to consider the
weaker executability $\Alive\imp\poss\tease\top$ instead of $\poss\tease\top$ in
$\EXE{\tease}{}$.

  On the other hand, implicit static laws may also indicate that the
inexecutability laws are too strong:

\begin{example}\label{ex_imp_stat3}
Consider $\STAT{}{}=\emptyset$,
$\EFF{\shoot}{}=\{\Loaded\imp\nec\shoot\neg\Alive\}$, 
$\EXE{\shoot}{}=\{\HasGun\imp\poss\shoot\top\}$ and
$\INEXE{\shoot}{}=\{\nec\shoot\bot\}$, with $\depbin$ still as above. For this
theory Algorithm~\ref{alg_imp_stat} returns $\IMPSTATtotal=\{\neg\HasGun\}$.
\end{example}

  In Example~\ref{ex_imp_stat3} we discovered that the agent never has a
gun. The problem here can be overcome by weakening $\nec\shoot\bot$ in
$\INEXE{\shoot}{}$ with $\neg\HasGun\imp\nec\shoot\bot$.\footnote{Regarding
Examples~\ref{ex_imp_stat2} and~\ref{ex_imp_stat3}, one might argue that in
practice such silly errors will never be made. Nevertheless, the examples here
given are quite simplistic, and for applications of real interest, whose
complexity will be much higher, we simply cannot rely on the designer's
knowledge about all side effects the stated formulas can have.}

  We can go further on this reasoning and also argue that the problem may be due
to a too strong set of effect laws or even to too strong frame axioms (\ie, a
too weak dependence relation). To witness, for Example~\ref{ex_imp_stat2}, if we
replace the law $\nec\tease\Walking$ by the weaker
$\Alive\imp\nec\tease\Walking$, the resulting action theory would satisfy
Postulate~\PostulStatLaw. In the same way, stating the (unintuitive) dependence
$\tease\depbin\Alive$ (which means the frame axiom
$\neg\Alive\imp\nec\tease\neg\Alive$ is no longer valid) guarantees satisfaction
of~\PostulStatLaw. (Note, however, that this solution becomes intuitive when
\Alive\ is replaced by \Awake.)

  To finish, implicit static laws of course may also indicate that the static
laws are too weak:

\begin{example}\label{ex_imp_stat1}
Suppose a computer representation of the line of integers, in which we can be at
a strictly positive number, $\Positive$, or at a negative one or zero,
$\neg\Positive$. Let $\MaxInt$ and $\MinInt$, respectively, be the largest and
the smallest representable integer number. $\goleft$ is the action of moving to
the biggest integer strictly smaller than the one at which we are. Consider the
following action theory for this scenario ($\At_{i}$ means we are at
number~$i$):
\[
\STAT{}{}=\{\At_{i}\imp\Positive : 0<i\leq\MaxInt\}\cup
        \{\At_{i}\imp\neg\Positive : \MinInt\leq i\leq 0\}
\]\[
\EFF{}{}= \begin{array}{c}
          \{\At_{\MinInt}\imp\nec\goleft\Underflow\}\cup \\
          \{\At_{i}\imp\nec\goleft\At_{i-1}: i>\MinInt\},\text{ }
       \end{array}
\EXE{}{}=\{\poss\goleft\top\},\ \INEXE{}{}=\emptyset
\]
with the dependence relation ($\MinInt\leq i\leq\MaxInt$):\\
\[
\depbin\ = \left\{
       \begin{array}{c}
          \depbintuple{\goleft,\At_{i}}, \depbintuple{\goleft,\Positive},\\
          \depbintuple{\goleft,\neg\Positive}, \depbintuple{\goleft,\Underflow}
       \end{array}
       \right\}
\]
Applying Algorithm~\ref{alg_imp_stat} to this action theory gives us all the
implicit static laws of the form $\neg(\At_{i}\et\At_{j})$, $i\neq j$, \ie, we
cannot be at two different numbers at the same~time.
\end{example}

  To summarize, in order to satisfy Postulate~\PostulStatLaw, an action theory
should contain a complete set of static laws or, alternatively, should not
contain too strong action laws.

\begin{remark}
$\STAT{}{}\cup\IMPSTATtotal$ in general is not intuitive.
\end{remark}

  Whereas in the latter example the implicit static laws should be added to
$\STAT{}{}$, in the others the implicit static laws are unintuitive and due to
an (in)executability law that is too strong and should be weakened. Of course,
how intuitive the modified action theory will be depends mainly on the knowledge
engineer's choice.

  To sum it up, eliminating implicit static laws may require revision of
$\STAT{}{}$, $\EFF{\act}{}$ or $\depbin$, or completion of $\EXE{\act}{}$ and
$\INEXE{\act}{}$. Completing $\INEXE{\act}{}$ is the topic we address in the
next section.

\section{No implicit inexecutability laws}\label{no_inexec}

  Let $\STAT{}{2}=\STAT{}{1}$, $\EFF{}{2}=\EFF{}{1}$ and $\INEXE{}{2}=\emptyset$
(executabilities do not matter here), and let $\depbin$ be that for
$\ActTheory{}{1}$. Note that $\ActTheory{}{2}$ satisfies
Postulate~\PostulStatLaw. From $\nec\tease\Walking$ it follows with $\STAT{}{2}$
that $\nec\tease\Alive$, \ie, in every situation, after teasing the turkey, it
is alive: $\STAT{}{2},\EFF{\tease}{2}\PDLvalid\nec\tease\Alive$. Now as
$\tease\indepbin\Alive$, the status of $\Alive$ is not modified by $\tease$, and
we have
$\STAT{}{2},\EFF{\tease}{2}\depbinvalid\neg\Alive\imp\nec\tease\neg\Alive$. From
the above, it follows
\[
\STAT{}{2},\EFF{\tease}{2},\EXE{\tease}{2},\INEXE{\tease}{2}
\depbinvalid\neg\Alive\imp\nec\tease\bot,
\]
\ie, an inexecutability law stating that a dead turkey cannot be teased. But
\[
\STAT{}{2},\INEXE{\tease}{2}\notPDLvalid\neg\Alive\imp\nec\tease\bot,
\]hence Postulate~\PostulInexeLaw\ is violated. Here the formula
$\neg\Alive\imp\nec\tease\bot$ is an example of what we call an {\em implicit
inexecutability~law}.

  In the literature, such laws are also known as {\em implicit
qualifications}~\cite{GinsbergSmith88}, and it has been often supposed, in a
more or less tacit way, that it is a positive feature of frameworks to leave
them implicit and provide mechanisms for inferring
them~\cite{Lin95,Thielscher97}. The other way round, one might argue as well
that implicit qualifications indicate that the domain has not been described in 
an adequate manner: the form of inexecutability laws is simpler than that of
effect laws, and it might be reasonably expected that it is easier to
exhaustively describe them.\footnote{Note that this concerns the necessary
conditions for executability, and thus it is not related to the qualification
problem, which basically says that it is difficult to state all the sufficient
conditions for executability.} Thus, all inexecutabilities of a given action
should be explicitly stated, and this is what Postulate~\PostulInexeLaw~says.

\myskip

  How can we check whether~\PostulInexeLaw\ is violated? We can conceive an
algorithm to find implicit inexecutability laws of a given action $\act$. The
basic idea is as follows: for every combination of effect laws of the form
$(\antec_{1}\et\ldots\et\antec_{n})\imp\nec\act(\conseq_{1}\et\ldots\et
\conseq_{n})$, with each $\antec_{i}\imp\nec\act\conseq_{i}\in\EFF{\act}{}$, if
$\antec_{1}\et\ldots\et\antec_{n}$ is consistent \wrt\ to \STAT{}{},
$\conseq_{1}\et\ldots\et\conseq_{n}$ inconsistent \wrt\ \STAT{}{}, and
$\STAT{}{},\INEXE{\act}{}\notPDLvalid(\antec_{1}\et\ldots\et\antec_{n})\imp
\nec\act\bot$, then output $(\antec_{1}\et\ldots\et\antec_{n})\imp\nec\act\bot$
as an implicit inexecutability law. Our algorithm basically does this, and
moreover takes into account dependence information.

  For an example of the execution of the algorithm, take $\ActTheory{\tease}{2}$
with $\depbin$ as given above. From $\EFF{\tease}{2}$ we get
$\top\imp\nec\tease\Walking$, whose antecedent is consistent with \STAT{}{}. As
$\depbinvalid\neg\Alive\imp\nec\tease\neg\Alive$ and $\STAT{}{}\cup\{\Walking\}
\models\Alive$, and because $\STAT{}{},\INEXE{\tease}{2}\notPDLvalid(\top\et
\neg\Alive)\imp\nec\tease\bot$, we caught an implicit inexecutability. As there
is no other combination of effect laws for \tease, we end the simulation here.

  Below is the pseudo-code of the algorithm for that (the reason $\EXE{\act}{}$
is not needed in the input will be made clear in the sequel):

\begin{algorithm}[Finding implicit inexecutability laws for $\act$]\mbox{ }\\
\label{alg_imp_inexec}
\vspace*{-0.35cm}
\begin{algorithmic}
\INPUT $\tuple{\STAT{}{},\EFF{\act}{},\INEXE{\act}{}}$ and $\depbin$
\OUTPUT $\IMPINEXE{\act}{}$, the set of implicit inexecutability laws for $\act$
\STATE $\IMPINEXE{\act}{}\GETS\ \emptyset$
\FORALL{$\hat{\EFF{\act}{}}\subseteq\EFF{\act}{}$}
   \STATE $\antec_{\hat{\EFF{\act}{}}}\GETS\ \bigwedge\{\antec_i :
           \antec_{i}\imp\nec\act\conseq_{i} \in \hat{\EFF{\act}{}}\}$ 
   \STATE $\conseq_{\hat{\EFF{\act}{}}}\GETS\ \bigwedge\{\conseq_i :
           \antec_{i}\imp\nec\act\conseq_{i} \in \hat{\EFF{\act}{}}\}$ 
   \FORALL{$\clause\in\NewCons{\STAT{}{}}{\conseq_{\hat{\EFF{\act}{}}}}$}
      \IF{$\forall\lit_{i}\in\clause,\ \act\indepbin\lit_{i}$ \ALGAND\
          $\STAT{}{},\INEXE{\act}{}\nvdash
           (\antec_{\hat{\EFF{\act}{}}}\et\neg\clause)\imp\nec\act\bot$}
         \STATE $\IMPINEXE{\act}{}\GETS\ \IMPINEXE{\act}{}\cup
                 \{(\antec_{\hat{\EFF{\act}{}}}\et\neg\clause)\imp\nec\act\bot\}$
      \ENDIF
   \ENDFOR
\ENDFOR
\end{algorithmic}
\end{algorithm}

\begin{theorem}~\label{theorem_imp_inexe_termination}
Algorithm~\ref{alg_imp_inexec} terminates.
\end{theorem}
\begin{proof}
Straightforward, as we have assumed $\STAT{}{}$, $\EFF{}{}$, $\INEXE{}{}$ and
$\depbin$ finite, and $\NewCons{}{}$ is finite (because $\STAT{}{}$ and
$\conseq_{\hat{\EFF{\act}{}}}$ are finite).
\end{proof}

\begin{example}\label{ex_imp_inexe}
Consider $\STAT{}{2},\EFF{\tease}{2},\INEXE{\tease}{2}$ and $\depbin$ as given
above. Then Algorithm~\ref{alg_imp_inexec} returns
$\IMPINEXE{\tease}{}=\{\neg\Alive\imp\nec\tease\bot\}$.
\end{example}

  Nevertheless, to apply Algorithm~\ref{alg_imp_inexec} is not enough to
guarantee Postulate~\PostulInexeLaw, as illustrated by the following example:

\begin{example}[Incompleteness of Algorithm~\ref{alg_imp_inexec}
without~\PostulStatLaw]
\label{ex_imp_inexe2}
Let $\STAT{}{}=\emptyset$, $\EFF{\act}{}=\{\prp_{1}\imp\nec\act\prp_{2}\}$,
$\EXE{\act}{}=\{\poss\act\top\}$, $\INEXE{\act}{}=\{\prp_{2}\imp\nec\act\bot\}$,
and $\depbin=\emptyset$. Then we have $\STAT{}{},\EFF{\act}{},\EXE{\act}{},
\INEXE{\act}{}\depbinvalid\prp_{1}\imp\nec\act\bot$, but after running
Algorithm~\ref{alg_imp_inexec} on $\ActTheory{\act}{}$ we have
$\STAT{}{},\IMPINEXE{\act}{}\notPDLvalid\prp_{1}\imp\nec\act\bot$.
\end{example}

  Example~\ref{ex_imp_inexe2} shows that the presence of implicit static laws
(induced by executabilities) implies the existence of implicit inexecutabilities
that are not caught by Algorithm~\ref{alg_imp_inexec}. One possibility of
getting rid of this is by considering the weaker version of~\PostulInexeLaw:

\begin{itemize}
\item[] \PostulInexeLaw' ({\bf No implicit inexecutability laws -- weak
version}):\\
\centerline{if $\STAT{}{},\EFF{\act}{},\EXE{\act}{},\INEXE{\act}{}\depbinvalid
\antec\imp\nec\act\bot$,  and $\STAT{}{},\EFF{\act}{},\EXE{\act}{},
\INEXE{\act}{}\notdepbinvalid\neg\antec$,}\\

\centerline{then $\STAT{}{},\INEXE{\act}{}\PDLvalid\antec\imp\nec\act\bot$}
\end{itemize}
If a non-trivial inexecutability law for a given action $\act$ can be inferred
from its respective theory, then it should be inferable in \PDL\ from the static
and inexecutability laws for it alone.

  With an adaptation of Algorithm~\ref{alg_imp_inexec} to take $\EXE{\act}{}$ in
its input and support a test for satisfiability of an inexecutability's
antecedent, we could guarantee completeness with respect to
Postulate~\PostulInexeLaw'. However such a test has the same complexity of
checking whether Postulate~\PostulStatLaw\ is satisfied. That is the reason we
keep abide on~\PostulInexeLaw\ and require $\ActTheory{\act}{}$ to satisfy
Postulate~\PostulStatLaw\ prior to running Algorithm~\ref{alg_imp_inexec}. This
gives us the following result:

\begin{theorem}\label{th_imp_inexe}
If $\ActTheory{\act}{}$ with $\depbin$ satisfies Postulate~\PostulStatLaw, then
it satisfies Postulate~\PostulInexeLaw\ if and only if $\IMPINEXE{\act}{}=
\emptyset$.
\end{theorem}
\begin{proof}
See Appendix~\ref{proof_th_imp_inexe}.
\end{proof}

\myskip

  With Algorithm~\ref{alg_imp_inexec}, not only do we decide whether
Postulate~\PostulInexeLaw\ is satisfied, but we also get information on how to
``repair'' the action theory. The set of implicit inexecutabilities so obtained
provides logical and metalogical information concerning the correction that must
be carried out: in the first case, elements of $\IMPINEXE{\act}{}$ can be added
to $\INEXE{\act}{}$; in the second one, $\IMPINEXE{\act}{}$ helps in properly
changing $\EFF{\act}{}$ or $\depbin$. For instance, to correct the action theory
of our example, the knowledge engineer would have the following options:
\begin{enumerate}
\item Add the qualification $\neg\Alive\imp\nec\tease\bot$ to
$\INEXE{\tease}{2}$; or
\item Add the (unintuitive) dependence $\depbintuple{\tease,\Alive}$ to
$\depbin$; or
\item Weaken the effect law $\nec\tease\Walking$ to
$\Alive\imp\nec\tease\Walking$ in $\EFF{\tease}{}$. 
\end{enumerate}
It is easy to see that whatever she opts for, the resulting action theory for
\tease\ will satisfy Postulate~\PostulInexeLaw\ (while still
satisfying~\PostulStatLaw).

\begin{example}[Drinking coffee~\cite{HerzigVarzinczak-IJCAI05}]
\label{coffee_example} 
Suppose, for instance, a hypothetical situation in which we reason about the
effects of drinking a cup of coffee:
\[
\STAT{}{}=\emptyset,\text{ }
\EFF{\drink}{} = \left\{
    \begin{array}{l}
        \Sugar\imp\nec\drink\Happy, \\
        \Salt\imp\nec\drink\neg\Happy
       \end{array}
     \right\},\text{ }
\EXE{\drink}{}=\INEXE{\drink}{}=\emptyset
\]and the dependence relation
\[
\depbin=\{\depbintuple{\drink,\Happy},\depbintuple{\drink,\neg\Happy}\}
\]Observe that $\ActTheory{\drink}{}$ satisfies~\PostulStatLaw. Then,
running Algorithm~\ref{alg_imp_inexec} on this action theory will give us
$\IMPINEXE{\drink}{}=\{(\Sugar\et\Salt)\imp\nec\drink\bot\}$.
\end{example}

\begin{remark}
$\INEXE{\act}{}\cup\IMPINEXE{\act}{}$ is not always intuitive.
\end{remark}

  Whereas in Example~\ref{ex_imp_inexe} we have got an inexecutability that
could be safely added to $\INEXE{\tease}{2}$, in Example~\ref{coffee_example} we
got an inexecutability that is unintuitive (just the presence of sugar and salt
in the coffee precludes drinking it). In that case, revision of other parts of
the theory should be considered in order to make it intuitive. Anyway, the
problem pointed out in the depicted scenario just illustrates that intuition is
beyond syntax. The scope of this work relies on the syntactical level. Only the
knowledge engineer can judge about how intuitive a formula is.

\myskip

  In what follows we revisit our postulates in order to strengthen them to the
case where more than one action is under concern and thus get results that can
be applied to whole action theories. 

\section{Generalizing the postulates}\label{generalizing}

  We have seen the importance that satisfaction of Postulates~\PostulConsist,
\PostulStatLaw\ and \PostulInexeLaw\ may have in describing the action theory of
a particular action $\act$. However, in applications of real interest more than
one action is involved, and thus a natural question that could be raised is
``can we have similar metatheoretical results for complex action theories''?

  In this section we generalize our set of postulates to action theories as a
whole, \ie, considering all actions of a domain, and prove some interesting
results that follow from that. As we are going to see, some of these results are
straightforward, while others must rely on some additional assumptions in order
to hold.

\myskip

  A generalization of Postulate~\PostulConsist\ is quite easy and has no need
for justification:

\begin{itemize}
\item[] \PostulConsistGen\ ({\bf Logical consistency}):
$\STAT{}{},\EFF{}{},\EXE{}{},\INEXE{}{}\notdepbinvalid\bot$
\end{itemize}
The whole action theory should be logically consistent.


\myskip

  Generalizing Postulate~\PostulStatLaw\ will give us the following:

\begin{itemize}
\item[] \PostulStatLawGen\ ({\bf No implicit static laws}):
\[
\mbox{if }\ 
\STAT{}{},\EFF{}{},\EXE{}{},\INEXE{}{}\depbinvalid\fml, \mbox{ then }\
\STAT{}{}\PDLvalid\fml
\]
\end{itemize}
If a classical formula can be inferred from the whole action theory, then it
should be inferable from the set of static laws alone. We have the following
results:

\begin{theorem}
$\ActTheory{}{}$ satisfies~\PostulStatLawGen\ if and only if
$\ActTheory{\act}{}$ satisfies~\PostulStatLaw\ for all $\act\in\Act$.
\end{theorem}
\begin{proof}

\noindent($\Rightarrow$): Straightforward.

\noindent($\Leftarrow$): Suppose $\ActTheory{}{}$ does not
satisfy~\PostulStatLawGen. Then there is $\fml\in\Fml$ such that
$\STAT{}{},\EFF{}{},\EXE{}{},\INEXE{}{}\depbinvalid\fml$ and
$\STAT{}{}\not\models\fml$. $\fml$ is equivalent to
$\fml_{1}\et\ldots\et\fml_{n}$, with $\fml_{1},\ldots,\fml_{n}\in\Fml$ and such
that there is at least one $\fml_{i}$ such that $\STAT{}{}\not\models\fml_{i}$
(otherwise $\STAT{}{}\models\fml$). Because the logic is independently
axiomatized, there must be some $\act\in\Act$ such that $\STAT{}{},\EFF{\act}{},
\EXE{\act}{},\INEXE{\act}{}\depbinvalid\fml_{i}$. From this and $\STAT{}{}\not
\models\fml_{i}$ it follows that $\ActTheory{\act}{}$ does not
satisfy~\PostulStatLaw.
\end{proof}

\begin{theorem}
If $\ActTheory{}{}$ satisfies~\PostulStatLawGen, then $\ActTheory{}{}$
satisfies~\PostulConsistGen\ if and only if $\ActTheory{\act}{}$
satisfies~\PostulConsist\ for all $\act\in\Act$.
\end{theorem}
\begin{proof}
Straightforward as the underlying logic is independently axiomatized.
\end{proof}

\comment{
  Under~\PostulStatLawGen\ we guarantee the converse of
Theorem~\ref{th_gen_completeness_exec_laws}:

\begin{theorem}\label{converse_th_gen_completeness_exec_laws}
Let $\ActTheory{}{}$ satisfy~\PostulStatLawGen. If $\ActTheory{}{}$
satisfies~\PostulCompleteExe*, then $\ActTheory{\act}{}$
satisfies~\PostulCompleteExe\ for all $\act\in\Act$.
\end{theorem}
\begin{proof}
Voir théorème~\ref{th_imp_exe_preserv}
\end{proof}
} 

  A more general form of Postulate \PostulInexeLaw\ can also be stated:

\begin{itemize}
\item[] \PostulInexeLawGen\ ({\bf No implicit inexecutability laws}):
\[
\mbox{if }\
\STAT{}{},\EFF{}{},\EXE{}{},\INEXE{}{}\depbinvalid\antec\imp\nec\act\bot,
\mbox{ then }\
\STAT{}{},\INEXE{}{}\PDLvalid\antec\imp\nec\act\bot
\]
\end{itemize}
If an inexecutability law can be inferred from the whole action theory, then it
should be inferable in \PDL\ from the static and inexecutability laws alone.

Note that having that $\ActTheory{\act}{}$ satisfies~\PostulInexeLaw\ for all
$\act\in\Act$ is not enough to $\ActTheory{}{}$ satisfy~\PostulInexeLawGen\ if
there are implicit static laws. To witness, let $\STAT{}{}=\EFF{\act_{1}}{}=
\emptyset$, and $\EXE{\act_{1}}{}=\{\poss{\act_{1}}\top\}$, $\INEXE{\act_{1}}{}=
\{\antec\imp\nec{\act_{1}}\bot\}$. Let also $\EFF{\act_{2}}{}=\EXE{\act_{2}}{}=
\INEXE{\act_{2}}{}=\emptyset$. Observe that both $\ActTheory{\act_{1}}{}$ and
$\ActTheory{\act_{2}}{}$ satisfy~\PostulInexeLaw, but
$\STAT{}{},\EFF{}{},\EXE{}{},\INEXE{}{}\depbinvalid\antec\imp\nec{\act_{2}}\bot$
and $\STAT{}{},\INEXE{}{}\notPDLvalid\antec\imp\nec{\act_{2}}\bot$.

  Nevertheless, under~\PostulStatLawGen\ the result follows:

\begin{theorem}
Let $\ActTheory{}{}$ satisfy~\PostulStatLawGen. $\ActTheory{}{}$
satisfies~\PostulInexeLawGen\ if and only if $\ActTheory{\act}{}$
satisfies~\PostulInexeLaw\ for all $\act\in\Act$.
\end{theorem}
\begin{proof}

\noindent($\Rightarrow$): Suppose that $\STAT{}{},\EFF{\act}{},\EXE{\act}{},
\INEXE{\act}{}\depbinvalid\fml\imp\nec\act\bot$. By monotonicity of
$\depbinvalid$, $\STAT{}{},\EFF{}{},\EXE{}{},\INEXE{}{}\depbinvalid\fml\imp\nec
\act\bot$, too. As $\ActTheory{\act}{}$ satisfies~\PostulInexeLawGen,
$\STAT{}{},\INEXE{}{}\PDLvalid\fml\imp\nec\act\bot$.

Now suppose that
$\STAT{}{},\INEXE{\act}{}\notPDLvalid\fml\imp\nec\act\bot$. Then there exists a
possible worlds model $\model=\tuple{\Worlds,\ract,\Val}$ such that
$\Mvalid{\model}\STAT{}{}\et\INEXE{\act}{}$ and there is a possible world
$v\in\Worlds$ such that $\wMvalid{v}{\model}\fml$ and $\notwMvalid{v}{\model}
\nec\act\bot$. Let $\model'=\tuple{\Worlds',\AccRel',\Val'}$ be such that
$\Worlds'=\Worlds$, $\Val'=\Val$, $\AccRel'_{\act'}=\emptyset$, for
$\act'\neq\act$, and $\AccRel'_{\act}=\ract$. Then $\Mvalid{\model'}\STAT{}{}\et
\INEXE{}{}$, and as $\STAT{}{},\INEXE{}{}\PDLvalid\fml\imp\nec\act\bot$, we get
a contradiction.


\noindent($\Leftarrow$): Suppose that $\ActTheory{}{}$ does not
satisfy~\PostulInexeLawGen. Then there exists $\fml\in\Fml$ such that
$\STAT{}{},\EFF{}{},\EXE{}{},\INEXE{}{}\depbinvalid\fml\imp\nec\act\bot$ and
$\STAT{}{},\INEXE{}{}\notPDLvalid\fml\imp\nec\act\bot$.

\noindent{\bf Claim:} $\STAT{}{},\EFF{\act}{},\EXE{\act}{},\INEXE{\act}{}
\depbinvalid\fml\imp\nec\act\bot$.

To witness, suppose $\STAT{}{},\EFF{\act}{},\EXE{\act}{},\INEXE{\act}{}
\notdepbinvalid\fml\imp\nec\act\bot$. Then there exists a possible worlds model
$\model=\tuple{\Worlds,\ract,\Val}$ such that
$\Mvalid{\model}\STAT{}{}\et\EFF{\act}{}\et\EXE{\act}{}\et\INEXE{\act}{}$ and
there is a possible world $v\in\Worlds$ such that $\wMvalid{v}{\model}\fml$ and
$\notwMvalid{v}{\model}\nec\act\bot$, \ie, there is $v'\in\Worlds$ such that
$\ract(v)=v'$. (We are going to extend $\model$ to be a model of
$\ActTheory{}{}$.)

For each $\act'\in\Act$, $\act'\neq\act$, we define:\\


\centerline{$\EFF{\act'}{\conseq}(w)=\{\conseq: \antec\imp\nec{\act'}\conseq
\in\EFF{\act'}{}\text{ and } \wMvalid{w}{\model}\conseq\}$}

\centerline{$\INEXE{\act'}{}(w)=\{\antec: \antec\imp\nec{\act'}\bot\in
\INEXE{\act'}{}\text{ and } \wMvalid{w}{\model}\antec\}$}

\centerline{$\EXE{\act'}{}(w)=\{\antec: \antec\imp\poss{\act'}\top\in
\EXE{\act'}{}\text{ and } \wMvalid{w}{\model}\antec\}$}

Let $\model'=\tuple{\Worlds',\AccRel',\Val'}$ be such that $\Worlds'=\Worlds$,
$\AccRel'=\ract\cup\bigcup_{\act'\neq\act}\AccRel_{\act'}$, and $\Val'=\Val$,
where for each $\act'$ and every world $w\in\Worlds'$:
\begin{itemize}
\item $\AccRel'(w)=\emptyset$, if $\INEXE{\act'}{}(w)\neq\emptyset$;
\item $\AccRel'(w)=w'$, if $\EFF{\act'}{\conseq}(w')\neq\emptyset$.
\end{itemize}
Because, by hypothesis, $\ActTheory{}{}$ satisfies~\PostulStatLawGen, there is
no implicit static law, \ie, $\STAT{}{}$ is complete in our sense. Then,
$\model'$ is a model of $\STAT{}{}$. We have that $\model'$ is a model of
$\EFF{}{}$, too: for every $\antec\imp\nec\act\conseq\in\EFF{}{}$ and every
$w\in\Worlds'$, if $\wMvalid{w}{\model'}\antec$, then $\wMvalid{w'}{\model'}
\conseq$ for all $w'\in\Worlds'$ such that $w\AccRel'w'$. Clearly $\model'$ is
also a model of $\INEXE{}{}$. $\model'$ is a model of $\EXE{}{}$, too: it is a
model of $\EXE{\act}{}$ and for every $\act'\neq\act$ and all those worlds
$w\in\Worlds'$ such that $\EXE{\act'}{}(w)\neq\emptyset$ there is a world
accessible by $\AccRel'$, \viz\ some $w'$ such that
$\EFF{\act'}{\conseq}(w')\neq\emptyset$ (because $\AccRel'(w)=\emptyset$ in this
case would preclude $\EXE{\act'}{}(w)\neq\emptyset$, as long
as~\PostulStatLawGen\ is satisfied). Thus $\Mvalid{\model'}\STAT{}{}
\et\EFF{}{}\et\EXE{}{}\et\INEXE{}{}$, but if this is the case,
$\STAT{}{},\EFF{}{},\EXE{}{},\INEXE{}{}\notdepbinvalid\fml\imp\nec\act\bot$,
hence we must have
$\STAT{}{},\EFF{\act}{},\EXE{\act}{},\INEXE{\act}{}\depbinvalid
\fml\imp\nec\act\bot$. (End of the proof of the claim.)

From $\STAT{}{},\INEXE{}{}\notPDLvalid\fml\imp\nec\act\bot$ it follows
$\STAT{}{},\INEXE{\act}{}\notPDLvalid\fml\imp\nec\act\bot$. Putting all the
results together, we have that $\ActTheory{\act}{}$ does not satisfy
Postulate~\PostulInexeLaw.
\end{proof}

  In the next section we make a step toward an attempt of amending our
modularity criteria by investigating possible extensions of our set of
postulates.

\section{Disturbing modularity}\label{disturbing}

  Can we augment our set of postulates to take into account other modules of
action theories or even other metatheoretical issues in reasoning about actions?
That is the topic we discuss in what follows.

\subsection{Postulates about effects of actions}

  It seems to be in line with our postulates to require action theories not to
allow for the deduction of new effect laws: if an effect law can be inferred
from an action theory (and no inexecutability for the same action in the same
context can be derived), then it should be inferable from the set of static and
effect laws alone. This means we should have:
\begin{itemize}
\item[] \PostulEffLaw\ ({\bf No implicit effect laws}):\\
\centerline{if
$\STAT{}{},\EFF{}{},\EXE{}{},\INEXE{}{}\depbinvalid\antec\imp\nec\act\conseq$
and
$\STAT{}{},\EFF{}{},\EXE{}{},\INEXE{}{}\notdepbinvalid
\antec\imp\nec\act\bot$,}\\
\centerline{then $\STAT{}{},\EFF{}{}\depbinvalid\antec\imp\nec\act\conseq$}
\end{itemize}
But consider the following intuitively correct action theory:
\[
\STAT{}{4}=\emptyset,\text{ }
\EFF{}{4}=\left\{
       \begin{array}{c}
          \Loaded\imp\nec\shoot\neg\Alive,\\
          (\neg\Loaded\et\Alive)\imp\nec\shoot\Alive
       \end{array}
     \right\}
\]\[
\EXE{}{4}=\{\HasGun\imp\poss\shoot\top\},\text{ }
\INEXE{}{4}=\{\neg\HasGun\imp\nec\shoot\bot\}
\]
together with the dependence $\shoot\depbin\neg\Alive$. It satisfies
Postulates~\PostulStatLawGen\ and \PostulInexeLawGen, but does not
satisfy~\PostulEffLaw. Indeed:
\[
\STAT{}{4},\EFF{}{4},\EXE{}{4},\INEXE{}{4}\depbinvalid
\neg\HasGun\ou\Loaded\imp\nec\shoot\neg\Alive
\]
and
\[
\STAT{}{4},\EFF{}{4},\EXE{}{4},\INEXE{}{4}\notdepbinvalid
\neg\HasGun\ou\Loaded\imp\nec\shoot\bot,
\]
but
\[
\STAT{}{4},\EFF{}{4}\notdepbinvalid\neg\HasGun\ou\Loaded\imp\nec\shoot\neg\Alive
\] 
So, Postulate~\PostulEffLaw\ would not help us to deliver the goods.


  Another possibility of improving our modularity criteria could be:
\begin{itemize}
\item[] \PostulAnatainEff\ ({\bf No unattainable effects}):
\[
\mbox{if }
\antec\imp\nec\act\conseq\in\EFF{}{}, \mbox{ then }
\STAT{}{},\EFF{}{},\EXE{}{},\INEXE{}{}\notdepbinvalid\antec\imp\nec\act\bot
\]
\end{itemize}
This expresses that if we have explicitly stated an effect law for $\act$ in
some context, then there should be no inexecutability law for the same action in
the same context. It is straightforward to design an algorithm which checks
whether this postulate is satisfied. We do not investigate this further here,
but just observe that the slightly stronger version below leads to unintuitive
consequences:
\begin{itemize}
\item[] \PostulAnatainEff' ({\bf No unattainable effects -- strong version}):
\[
\mbox{if }
\STAT{}{},\EFF{}{}\depbinvalid\antec\imp\nec\act\conseq, \mbox{ then }
\STAT{}{},\EFF{}{},\EXE{}{},\INEXE{}{}\notdepbinvalid\antec\imp\nec\act\bot
\]
\end{itemize}
Indeed, for the above action theory we have 
\[
\EFF{}{4}\depbinvalid (\neg\HasGun\et\Loaded)\imp\nec\shoot\neg\Alive,
\]
but
\[
\STAT{}{4},\EFF{}{4},\EXE{}{4},\INEXE{}{4}\depbinvalid(\neg\HasGun\et\Loaded)
\imp\nec\shoot\bot.
\]
This is certainly too strong. Our example also illustrates that it is sometimes
natural to have `redundancies' or `overlaps' between $\EFF{}{}$
and~$\INEXE{}{}$. Indeed, as we have pointed out, inexecutability laws are a
particular kind of effect laws, and the distinction here made is
conventional. The decision of considering them as strictly different entities or
not depends mainly on the context. At a representational level we prefer to keep
them separated, while in Algorithm~\ref{alg_imp_stat} we have mixed them
together in order to compute the consequences of an action.

\medskip

  In what follows we address the problem of completing the set of executability
laws of an action theory.

\subsection{Maximizing executabilities}

  As we have seen, implicit static laws only show up when there are
executability laws. So, a question that naturally raises is ``which
executability laws can be consistently added to a given action theory?''.

  A hypothesis usually made in the literature is that of maximization of
executabilities: in the absence of a proof that an action is inexecutable in a
given context, assume its executability for that context. Such a hypothesis is
captured by the following postulate that we investigate in this section:

\begin{itemize}
\item[] \PostulCompleteExe\ ({\bf Maximal executability laws}):
\[
\mbox{if }\
\STAT{}{},\EFF{\act}{},\EXE{\act}{},\INEXE{\act}{}\notdepbinvalid\antec\imp
\nec\act\bot, \mbox{ then }\
\STAT{}{},\EXE{\act}{}\PDLvalid\antec\imp\poss\act\top
\]
\end{itemize}
Such a postulate expresses that if in context $\antec$ no inexecutability for
$\act$ can be inferred, then the respective executability should follow in \PDL\
from the executability and static laws.

  Postulate~\PostulCompleteExe\ generally holds in nonmonotonic frameworks, and
can be enforced in monotonic approaches such as ours by maximizing
$\EXE{\act}{}$. We nevertheless would like to point out that maximizing
executability is not always intuitive. To witness, suppose we know that if we
have the ignition key, the tank is full, $\ldots$, and the battery tension is
beyond 10V, then the car (necessarily) will start. Suppose we also know that if
the tension is below 8V, then the car will not start. What should we conclude in
situations where we know that the tension is 9V? Maximizing executabilities
makes us infer that it will start, but such reasoning is not what we want if we
would like to be sure that all possible executions lead to the goal.

\section{Exploiting modularity}\label{discussion}

  In this section we present other properties related to consistency and
modularity of action theories, emphasizing the main results that we obtain when
Postulate~\PostulStatLawGen\ is satisfied.

\begin{theorem}
If $\ActTheory{}{}$ satisfies Postulate~\PostulStatLawGen, then
$\STAT{}{},\EFF{}{},\EXE{}{},\INEXE{}{}\depbinvalid\bot$ if and only if
$\STAT{}{}\models\bot$.
\end{theorem}
This theorem says that if there are no implicit static laws, then consistency of
an action theory can be checked by just checking consistency of \STAT{}{}.

\begin{theorem}
\label{th_intra_mod_eff}
If $\ActTheory{}{}$ satisfies Postulate~\PostulStatLawGen, then
$\STAT{}{},\EFF{}{},\EXE{}{},\INEXE{}{}\depbinvalid\antec\imp\nec\act\conseq$ if
and only if
$\STAT{}{},\EFF{\act}{},\INEXE{\act}{}\depbinvalid\antec\imp\nec\act\conseq$.
\end{theorem}
\begin{proof}

\noindent($\Leftarrow$): Straightforward, by monotonicity.

\noindent($\Rightarrow$): Suppose that $\STAT{}{},\EFF{\act}{},\INEXE{\act}{}
\notdepbinvalid\antec\imp\nec\act\conseq$. Then there exists a possible worlds
model $\model\in\modelclass{\depbin}$, $\model=\tuple{\Worlds,\ract,\Val}$,
such that $\Mvalid{\model}\STAT{}{}\et\EFF{\act}{}\et\INEXE{\act}{}$ and there
is a possible world $v\in\Worlds$ such that $\wMvalid{v}{\model}\antec$ and
$\notwMvalid{v}{\model}\nec\act\conseq$, \ie, there is $v'\in\Worlds$ such that 
$\ract(v)=v'$ and $\notwMvalid{v'}{\model}\conseq$. (We are going to extend
$\model$ to obtain a model of $\ActTheory{}{}$ and thus show that $\STAT{}{},
\EFF{}{},\EXE{}{},\INEXE{}{}\notdepbinvalid\antec\imp\nec\act\conseq$.)

For each $\act'\in\Act$, $\act'\neq\act$, we define:\\


\centerline{$\EFF{\act'}{\conseq}(w)=\{\conseq: \antec\imp\nec{\act'}\conseq
\in\EFF{\act'}{}\text{ and } \wMvalid{w}{\model}\conseq\}$}

\centerline{$\INEXE{\act'}{}(w)=\{\antec: \antec\imp\nec{\act'}\bot\in
\INEXE{\act'}{}\text{ and } \wMvalid{w}{\model}\antec\}$}

\centerline{$\EXE{\act'}{}(w)=\{\antec: \antec\imp\poss{\act'}\top\in
\EXE{\act'}{}\text{ and } \wMvalid{w}{\model}\antec\}$}

Let $\model'=\tuple{\Worlds',\AccRel',\Val'}$ be such that $\Worlds'=\Worlds$,
$\AccRel'=\ract\cup\bigcup_{\act'\neq\act}\AccRel_{\act'}$, and $\Val'=\Val$,
where for each $\act'$ and every world $w\in\Worlds'$:
\begin{itemize}
\item $\AccRel'(w)=\emptyset$, if $\INEXE{\act'}{}(w)\neq\emptyset$;
\item $\AccRel'(w)=w'$, if $\EFF{\act'}{\conseq}(w')\neq\emptyset$.
\end{itemize}
Because, by hypothesis, $\ActTheory{}{}$ satisfies~\PostulStatLawGen, there is
no implicit static law, \ie, $\STAT{}{}$ is complete in our sense. Then,
$\model'$ is a model of $\STAT{}{}$. We have that $\model'$ is a model of
$\EFF{}{}$, too: for every $\antec\imp\nec\act\conseq\in\EFF{}{}$ and every
$w\in\Worlds'$, if $\wMvalid{w}{\model'}\antec$, then $\wMvalid{w'}{\model'}
\conseq$ for all $w'\in\Worlds'$ such that $w\AccRel'w'$. Clearly $\model'$ is
also a model of~$\INEXE{}{}$. $\model'$ is a model of $\EXE{}{}$, too: it is a
model of $\EXE{\act}{}$ and for every $\act'\neq\act$ and all those worlds
$w\in\Worlds'$ such that $\EXE{\act'}{}(w)\neq\emptyset$ there is a world
accessible by $\AccRel'$, \viz\ some $w'$ such that
$\EFF{\act'}{\conseq}(w')\neq\emptyset$ (because $\AccRel'(w)=\emptyset$ in this
case would preclude $\EXE{\act'}{}(w)\neq\emptyset$, as long
as~\PostulStatLawGen\ is satisfied). Hence $\Mvalid{\model'}\STAT{}{}
\et\EFF{}{}\et\EXE{}{}\et\INEXE{}{}$. Because there are $v,v'\in\Worlds'$ such
that $\wMvalid{v}{\model'}\antec$, $\AccRel'(v)=v'$ and
$\notwMvalid{v'}{\model'}\conseq$, we have
$\STAT{}{},\EFF{}{},\EXE{}{},\INEXE{}{}\notdepbinvalid
\antec\imp\nec\act\conseq$.
\end{proof}

This means that under~\PostulStatLawGen\ we have modularity inside \EFF{}{},
too: when deducing the effects of $\act$ we need not consider the action laws
for other actions. Versions for executability and inexecutability can be stated
as well:

\begin{theorem}
\label{th_intra_mod_exe}
If $\ActTheory{}{}$ satisfies Postulate~\PostulStatLawGen, then
$\STAT{}{},\EFF{}{},\EXE{}{},\INEXE{}{}\depbinvalid\antec\imp\poss\act\top$ if
and only if $\STAT{}{},\EXE{\act}{}\depbinvalid\antec\imp\poss\act\top$.
\end{theorem}
\begin{proof}

\noindent($\Leftarrow$): Straightforward, by monotonicity.

\noindent($\Rightarrow$): Suppose that $\STAT{}{},\EXE{\act}{}\notdepbinvalid
\antec\imp\poss\act\top$. Then there exists a possible worlds model
$\model\in\modelclass{\depbin}$, $\model=\tuple{\Worlds,\ract,\Val}$, such that
$\Mvalid{\model}\STAT{}{}\et\EXE{\act}{}$ and there is a possible world
$v\in\Worlds$ such that $\wMvalid{v}{\model}\antec$ and
$\notwMvalid{v}{\model}\poss\act\top$. (We are going to extend $\model$ to build
a model of $\ActTheory{}{}$ and thus conclude that $\STAT{}{},\EFF{}{},
\EXE{}{},\INEXE{}{}\notdepbinvalid\antec\imp\poss\act\top$.)

For each $\act'\in\Act$, $\act'\neq\act$, we define:\\


\centerline{$\EFF{\act'}{\conseq}(w)=\{\conseq: \antec\imp\nec{\act'}\conseq
\in\EFF{\act'}{}\text{ and } \wMvalid{w}{\model}\conseq\}$}

\centerline{$\INEXE{\act'}{}(w)=\{\antec: \antec\imp\nec{\act'}\bot\in
\INEXE{\act'}{}\text{ and } \wMvalid{w}{\model}\antec\}$}

\centerline{$\EXE{\act'}{}(w)=\{\antec: \antec\imp\poss{\act'}\top\in
\EXE{\act'}{}\text{ and } \wMvalid{w}{\model}\antec\}$}

Let $\model'=\tuple{\Worlds',\AccRel',\Val'}$ be such that $\Worlds'=\Worlds$,
$\AccRel'=\ract\cup\bigcup_{\act'\neq\act}\AccRel_{\act'}$, and $\Val'=\Val$,
where for each $\act'$ and every world $w\in\Worlds'$:
\begin{itemize}
\item $\AccRel'(w)=\emptyset$, if $\INEXE{\act'}{}(w)\neq\emptyset$;
\item $\AccRel'(w)=w'$, if $\EFF{\act'}{\conseq}(w')\neq\emptyset$.
\end{itemize}
Because, by hypothesis, $\ActTheory{}{}$ satisfies~\PostulStatLawGen, there is
no implicit static law, \ie, $\STAT{}{}$ is complete in our sense. Then,
$\model'$ is a model of $\STAT{}{}$. We have that $\model'$ is a model of
$\EFF{}{}$, too: for every $\antec\imp\nec\act\conseq\in\EFF{}{}$ and every
$w\in\Worlds'$, if $\wMvalid{w}{\model'}\antec$, then $\wMvalid{w'}{\model'}
\conseq$ for all $w'\in\Worlds'$ such that $w\AccRel'w'$. Clearly $\model'$ is
also a model of $\INEXE{}{}$. $\model'$ is a model of $\EXE{}{}$, too: it is a
model of $\EXE{\act}{}$ and for every $\act'\neq\act$ and all those worlds
$w\in\Worlds'$ such that $\EXE{\act'}{}(w)\neq\emptyset$ there is a world
accessible by $\AccRel'$, \viz\ some $w'$ such that
$\EFF{\act'}{\conseq}(w')\neq\emptyset$ (because $\AccRel'(w)=\emptyset$ in this
case would preclude $\EXE{\act'}{}(w)\neq\emptyset$, as long
as~\PostulStatLawGen\ is satisfied). Hence $\Mvalid{\model'}\STAT{}{}
\et\EFF{}{}\et\EXE{}{}\et\INEXE{}{}$. Because there is $v\in\Worlds'$ such that
$\wMvalid{v}{\model'}\antec$  and $\notwMvalid{v}{\model'}\poss\act\top$, we
have
$\STAT{}{},\EFF{}{},\EXE{}{},\INEXE{}{}\notdepbinvalid\antec\imp\poss\act\top$.
\end{proof}

\begin{corollary}\label{PCconseqPS}
\PostulExeLaw\ is a consequence of \PostulStatLaw.
\end{corollary}
\begin{proof}
Straightforward.
\end{proof}

\begin{theorem}
\label{th_intra_mod_inexe}
If $\ActTheory{}{}$ satisfies Postulates~\PostulStatLawGen\
and~\PostulInexeLawGen, then $\STAT{}{},\EFF{}{},\EXE{}{},\INEXE{}{}\depbinvalid
\antec\imp\nec\act\bot$ if and only if $\STAT{}{},\INEXE{\act}{}\depbinvalid
\antec\imp\nec\act\bot$.
\end{theorem}
\begin{proof}

\noindent($\Leftarrow$): Straightforward, by monotonicity.

\noindent($\Rightarrow$): If $\STAT{}{},\EFF{}{},\EXE{}{},\INEXE{}{}\depbinvalid
\antec\imp\nec\act\bot$, then from~\PostulStatLawGen\ and
Theorem~\ref{th_intra_mod_eff} we have $\STAT{}{},\EFF{\act}{},\INEXE{\act}{}
\depbinvalid\antec\imp\nec\act\bot$. From this and~\PostulInexeLawGen\ it
follows $\STAT{}{},\INEXE{\act}{}\depbinvalid\antec\imp\nec\act\bot$.
\end{proof}

\comment{
\begin{theorem}
There exist action theories $\ActTheory{}{}$ not satisfying~\PostulStatLawGen\
such that
$\STAT{}{},\EFF{}{},\EXE{}{},\INEXE{}{}\depbinvalid\antec\imp\nec\act\conseq$ 
and
$\STAT{}{},\EFF{\act}{},\INEXE{\act}{}\notdepbinvalid\antec\imp\nec\act\conseq$.
\end{theorem}
For example, we have that
\[
\STAT{}{1},\EFF{}{1},\EXE{}{1},\INEXE{}{1}\depbinvalid\neg\Alive\imp
\nec\shoot\Alive,
\]but
\[
{\STAT{}{1}},\EFF{\shoot}{1},\INEXE{\shoot}{1}\notdepbinvalid
\neg\Alive\imp\nec\shoot\Alive.
\]
} 

\section{Related work}\label{rel_works}

  Pirri and Reiter have investigated the metatheory of the Situation
Calculus~\cite{PirriReiter99}. In a spirit similar to ours, they use
executability laws and effect laws. Contrarily to us, their executability laws
are equivalences and are thus at the same time inexecutability laws. As they
restrict themselves to domains without ramifications, there are no static laws, 
\ie, $\STAT{}{}=\emptyset$. For this setting they give a syntactical condition on
effect laws guaranteeing that they do not interact with the executability laws
in the sense that they do not entail implicit static laws. Basically, the
condition says that when there are effect laws $\antec_{1}\imp\nec\act\conseq$ 
and $\antec_{2}\imp\nec\act\neg\conseq$, then $\antec_{1}$ and $\antec_{2}$ are
inconsistent (which essentially amounts to having in their theories a kind of
``implicit static law schema'' of the form $\neg(\antec_{1}\et\antec_{2})$).

 This then allows them to show that such theories are always
consistent. Moreover they thus simplify the entailment problem for this
calculus, and show for several problems such as consistency or regression that
only some of the modules of an action theory are necessary.

\myskip

  Amir~\cite{Amir-AAAI2000} focuses on design and maintainability of action
descriptions applying many of the concepts of the object-oriented paradigm in
the Situation Calculus. In that work, guidelines for a partitioned
representation of a given theory are presented, with which the inference task
can also be optimized, as it is restricted to the part of the theory that is
really relevant to a given query. This is observed specially when different
agents are involved: the design of an agent's theory can be done with no regard
to others', and after the integration of multiple agents, queries about an
agent's beliefs do not take into account the belief state of other agents.

  In the referred work, executabilities are as in~\cite{PirriReiter99} and the
same condition on effect laws is assumed, which syntactically precludes the
existence of implicit static laws.

  Despite of using many of the object-oriented paradigm tools and techniques, no
mention is made to the concepts of cohesion and coupling~\cite{Pressman92},
which are closely related to modularity~\cite{HerzigVarzinczak-IJCAI05}. In the
approach presented in~\cite{Amir-AAAI2000}, even if modules are highly
cohesive, they are not necessarily lowly coupled, due to the dependence between
objects in the reasoning phase. We do not investigate this further here, but
conjecture that this could be done there by, during the reasoning process
defined for that approach, avoiding passing to a module a formula of a type
different from those it contains.

\myskip

  The present work generalizes and extends Pirri and Reiter's result to the case
where $\STAT{}{}\neq\emptyset$ and both these works where the syntactical
restriction on effect laws is not made. This gives us more expressive power, as
we can reason about inexecutabilities, and a better modularity in the sense that
we do not combine formulas that are conceptually different (\viz\
executabilities and inexecutabilities).

\myskip

  Zhang \etal~\cite{ZhangEtAl02} have also proposed an assessment of what a good
action theory should look like. They develop the ideas in the framework of
\EPDL~\cite{ZhangFoo01}, an extended version of \PDL\ which allows for
propositions as modalities to represent causal connection between literals. We
do not present the details of that, but concentrate on the main metatheoretical
results.

  Zhang \etal\ propose a normal form for describing action
theories,\footnote{But not as expressive as one might think: For instance, in 
modeling the nondeterministic action of dropping a coin on a chessboard, we are
not able to state $\nec\drop(\Black\ou\White)$. Instead, we should write
something like $\nec{\drop_{\Black}}\Black$, $\nec{\drop_{\White}}\White$,
$\nec{\drop_{\Black,\White}}\Black$ and $\nec{\drop_{\Black,\White}}\White$,
where $\drop_{\Black}$ is the action of dropping the coin on a black square 
(analogously for the others) and
$\drop=\drop_{\Black}\cup\drop_{\White}\cup\drop_{\Black,\White}$, with
``$\cup$'' the nondeterministic composition of actions.} and investigate three
levels of consistency. Roughly speaking, an action theory $\THE{}{}$ is {\em
uniformly consistent} if it is globally consistent (\ie,
$\THE{}{}\notEPDLvalid\bot$); a formula $\modfml$ is $\THE{}{}$-{\em consistent}
if $\THE{}{}\notEPDLvalid\neg\modfml$, for $\THE{}{}$ a uniformly consistent
theory; $\THE{}{}$ is {\em universally consistent} if (in our terms) every
logically possible world is accessible. $\THE{}{}\EPDLvalid\fml$ implies
$\EPDLvalid\fml$.

  Furthermore, two assumptions are made to preclude the existence of implicit
qualifications. Satisfaction of such assumptions means the action theory under
consideration is {\em safe}, \ie, it is uniformly consistent. Such a normal form
justifies the two assumptions made and on whose validity relies their notion of
good action theories.

  Given these definitions, they propose algorithms to test the different
versions of consistency for an action theory $\THE{}{}$ that is in normal 
form. This test essentially amounts to checking whether $\THE{}{}$ is {\em safe},
\ie, whether $\THE{}{}\EPDLvalid\poss\act\top$, for every action~$\act$. Success
of this check should mean the action theory under analysis satisfies the
consistency requirements.

  Nevertheless, this is only a necessary condition: it is not hard to imagine
action theories that are uniformly consistent but in which we can still have
implicit inexecutabilities that are not caught by their algorithm. Consider for
instance a scenario with a lamp that can be turned on and off by a toggle
action, and its \EPDL\ representation given by:
\[
\THE{}{}=\left\{
  \begin{array}{c}
    \On\imp\nec\toggle\neg\On, \\
    \Off\imp\nec\toggle\On, \\
    \nec\On\neg\Off, \\
    \nec{\neg\On}\Off
  \end{array}
\right\}
\]
The causal statement $\nec\On\neg\Off$ means that \On\ causes $\neg\Off$. Such
an action theory satisfies each of the consistency requirements (in particular
it is uniformly consistent, as $\THE{}{}\notEPDLvalid\bot$). Nevertheless,
$\THE{}{}$ is not safe for the static law $\neg(\On\et\Off)$ cannot be
proved.\footnote{A possible solution could be to consider the set of static
constraints explicitly in the action theory (viz.\ in the deductive system). For
the running example, taking into account the constraint $\On\sii\neg\Off$
(derived from the causal statements and the \EPDL\ global axioms), we can
conclude that $\THE{}{}$ is safe. On the other hand, all the side effects such a
modification could have on the whole theory has yet to be analyzed.}

  Although they are concerned with the same kind of problems that have been
discussed in this paper, they take an overall view of the subject, in the sense
that all problems are dealt with together. This means that in their approach no
special attention (in our sense) is given to the different components of the
action theory, and then every time something is wrong with it this is taken as a
global problem inherent to the action theory as a whole. Whereas such a
``systemic'' view of action theories is not necessarily a drawback (we have just
seen the strong interaction that exists between the different sets of laws
composing an action theory), being modular in our sense allows us to
better identify the ``problematic'' laws and take care of them. Moreover, the
advantage of allowing to find the set of laws which must be modified in order to
achieve the desired consistency is made evident by the algorithms we have
proposed (while their results only allow to decide whether a given theory
satisfies some consistency requirement).

\myskip

  Lang \etal~\cite{LangEtAl03} address consistency in the causal laws
approach~\cite{McCainTurner95}, focusing on the computational aspects. They
suppose an abstract notion of completion of an action theory solving the frame 
problem. Given an action theory $\THE{\act}{}$ containing logical information
about $\act$'s direct effects as well as the indirect effects that may follow
(expressed in the form of causal laws), the completion of $\THE{\act}{}$ roughly
speaking is the original theory $\THE{\act}{}$ amended of logical axioms stating
the persistence of all non-affected (directly nor indirectly) literals. (Note
that such a notion of completion is close to the underlying semantics of the
dependence relation used throughout the present paper, which essentially amounts
to the explanation closure assumption~\cite{Schubert90}.)

  Their \mboxsc{executability} problem is to check whether action $\act$ is
executable in all possible initial states (Zhang~\etal's safety property). This
amounts to testing whether every possible state $w$ has a successor $w'$
reachable by $\act$ such that $w$ and $w'$ both satisfy the completion of
$\THE{\act}{}$. For instance, still considering the lamp scenario, the
representation of the action theory for \toggle\ is:
\[
\THE{\toggle}{}=\left\{
  \begin{array}{c}
    \On\stackrel{\toggle}{\longrightarrow}\Off, \\
    \Off\stackrel{\toggle}{\longrightarrow}\On, \\
    \Off\longrightarrow\neg\On, \\
    \On\longrightarrow\neg\Off
  \end{array}
\right\}
\]
where the first two formulas are conditional effect laws for \toggle, and
the latter two causal laws in McCain and Turner's sense. We will not dive in the
technical details, and just note that the executability check will return ``no''
for this example as \toggle\ cannot be executed in a state satisfying
$\On\et\Off$.

  In the mentioned work, the authors are more concerned with the complexity
analysis of the problem of doing such a consistency test and no algorithm for 
performing it is given, however. In spite of the fact their motivation is the
same as ours, again what is presented is a kind of ``yes-no tool'' which can
help in doing a metatheoretical analysis of a given action theory, and many of
the comments concerning Zhang \etal's approach could be repeated here.

\myskip

  Another criticism that could be made about both these approaches concerns the
assumption of full executability they rely on. We find it too strong to require
all actions to be always executable, and to reject as bad an action theory
admitting situations where some action cannot be executed at all. As an example,
consider the very simple action theory given by $\STAT{}{5}=\STAT{}{1}$,
$\EFF{}{5}=\{\nec\tease\Walking\}$, $\EXE{}{5}=\EXE{}{1}$ and
$\INEXE{}{5}=\INEXE{}{1}$, and consider
$\depbin=\{\depbintuple{\tease,\Walking}\}$. Observe that, with our approach, it
suffices to derive the implicit inexecutability law
$\neg\Alive\imp\nec\tease\bot$, change $\INEXE{}{}$, and the system will properly
run in situations where $\neg\Alive$ is the case.

  On the other hand, if we consider the equivalent representation of such an
action theory in the approach of Lang~\etal, after computing the completion of
$\THE{\tease}{}$, if we test its executability, we will get the answer ``no'',
the reason being that \tease\ is not executable in the possible state where
$\neg\Alive$ holds. Such an answer is correct, but note that with only this as
guideline we have no idea about where a possible modification in the action
theory should be carried on in order to achieve full executability for
\tease. The same observation holds for Zhang \etal's proposal.

  Just to see how things can be even worse, consider the action theory
$\tuple{{\STAT{'}{5}},{\EFF{'}{5}},{\EXE{'}{5}},{\INEXE{'}{5}}}$, with
${\STAT{'}{5}}=\STAT{}{5}$, ${\EFF{'}{5}}=\EFF{}{5}$,
${\EXE{'}{5}}=\{\Alive\imp\poss\tease\top\}$ and
${\INEXE{'}{5}}=\{\neg\Alive\imp\nec\tease\bot\}$, with the same $\depbin$,
obtained by the correction of $\ActTheory{}{5}$ above with the algorithms we
propose. Observe that
$\tuple{{\STAT{'}{5}},{\EFF{'}{5}},{\EXE{'}{5}},{\INEXE{'}{5}}}$ satisfies all
our postulates. It is not hard to see, however, that the representation of such
an action theory in the above frameworks, when checked by their respective
consistency tests, is still considered to have a problem.

  This problem arises because Lang~\etal's proposal do not allow for
executability laws, thus one cannot make the distinction between
$\EXE{}{}=\{\poss\tease\top\}$, $\EXE{}{}=\{\Alive\imp\poss\tease\top\}$ and
$\EXE{}{}=\emptyset$. By their turn, Zhang \etal's allows for specifying
executabilities, but their consistency definitions do not distinguish the cases
$\Alive\imp\poss\tease\top$ and $\poss\tease\top$.

\myskip

  A concept similar to that of implicit static laws was firstly addressed, as
far as we are concerned, in the realm of regulation consistency with deontic
logic~\cite{Cholvy99}. Indeed, the notions of regulation consistency given in
the mentioned work and that of modularity presented
in~\cite{HerzigVarzinczak-Aiml04Proc05} and refined here can be proved to be
equivalent. The main difference between the mentioned work and the approach
in~\cite{HerzigVarzinczak-Aiml04Proc05} relies on the fact that
in~\cite{Cholvy99} some syntactical restrictions on the formulas have to be made
in order to make the algorithm to work.

\myskip

  Lifschitz and Ren~\cite{LifschitzRen2006} propose an action description
language derived from $\mathcal{C}+$~\cite{GiunchigliaEtAl-AIJ.2004} in which
domain descriptions can also be decomposed in modules. Contrarily to our
setting, in theirs a module is not a set of formulas for given action $\act$,
but rather a description of a subsystem of the theory, \ie, each module
describes a set of interrelated fluents and actions. As an example, a module
describing Lin's suitcase~\cite{Lin95} should contain all causal laws in the
sense of $\mathcal{C}+$ that are relevant to the scenario. Actions or fluents
having nothing to do, neither directly nor indirectly, with the suitcase should
be described in different modules. This feature makes such a decomposition
somewhat domain-dependent, while here we have proposed a type-oriented
modularization of the formulas, which does not depend on the domain.

  In the referred work, modules can be defined in order to specialize other
modules. This is done by making the new module to inherit and then specialize
other modules' components. This is an important feature when elaborations are
involved. In the suitcase example, adding a new action relevant to the suitcase
description can be achieved by defining a new module inheriting all properties
of the old one and containing the causal laws needed for the new action. Such
ideas are interesting from the standpoint of software and knowledge engineer:
reusability is an intrinsic property of the framework, and easy scalability
promotes elaboration tolerance.

  Consistency of a given theory and how to prevent conflicts between modules
(independent or inherited) however is not addressed.

\myskip

  In this work we have illustrated by some examples what we can do in order to
make a theory intuitive. This involves theory modification. Action theory change
has been addressed in the recent literature on revision and
update~\cite{LiPereira96,Liberatore-JELIA2000,EiterEtAl-IJCAI05}.
In~\cite{HerzigEtAl-TechRep06} we have investigated this issue and shown the
importance that modularity has in such a task.

\section{Conclusion}

  Our contribution is twofold: general, as we presented postulates that apply to
all reasoning about actions formalisms; and specific, as we proposed algorithms
for a dependence-based solution to the frame problem.

  We have defined here the concept of modularity of an action theory and pointed
out some of the problems that arise if it is not satisfied. In particular we
have argued that the non-dynamic part of action theories could influence but
should not be influenced by the dynamic~one.\footnote{It might be objected that
it is only by doing experiments that one learns the static laws that govern the
universe. But note that this involves {\em learning}, whereas here -- as always
done in the reasoning about actions field -- the static laws are known once
forever, and do not evolve.}

  We have put forward some postulates, and in particular tried to demonstrate
that when there are implicit static and inexecutability laws then one has
slipped up in designing the action theory in question. As shown, a possible
solution comes into its own with Algorithms~\ref{alg_imp_stat}
and~\ref{alg_imp_inexec}, which can give us some guidelines in correcting an
action theory~if~needed. By means of examples we have seen that there are
several alternatives of correction, and choosing the right module to be modified
as well as providing the intuitive information that must be supplied is up to
the knowledge engineer.

\myskip

  Given the difficulty of exhaustively enumerating all the preconditions under
which a given action is executable (and also those under which such an action
cannot be executed), it is reasonable to expect that there is always going to be
some executability precondition $\fml_{1}$ and some inexecutability
precondition~$\fml_{2}$ that together lead to a contradiction, forcing, thus, an
implicit static law $\neg(\fml_{1}\et\fml_{2})$. This is the reason we propose
to state some information about both executabilities and inexecutabilities, and
then run the algorithms in order to improve the description.

  It could be argued that unintuitive consequences in action theories are mainly
due to badly written axioms and not to the lack of modularity. True enough, but
what we have presented here is the case that making a domain description modular
gives us a tool to detect at least some of such problems and correct it. (But
note that we do not claim to correct badly written axioms automatically and once
for all.) Besides this, having separate entities in the ontology and controlling
their interaction help us to localize where the problems are, which can be
crucial for real world applications.

\myskip

  In this work we used a version of \PDL, but our notions and results can be
applied to other frameworks as well. It is worth noting however that for
first-order based frameworks the consistency checks of
Algorithms~\ref{alg_imp_stat} and~\ref{alg_imp_inexec} are undecidable. We can
get rid of this by assuming that $\ActTheory{}{}$ is finite and there is no
function symbol in the language. In this way, the result of \NewCons{}{} is
finite and the algorithm terminates.

  The present paper is also a step toward a solution to the problem of indirect
dependences: indeed, if the indirect dependence $\shoot\depbin\neg\Walking$ is
not in~$\depbin$, then after running Algorithm~\ref{alg_imp_inexec} we get an
indirect inexecutability $(\Loaded\et\Walking)\imp\nec\shoot\bot$, \ie, $\shoot$
cannot be executed if $\Loaded\et\Walking$ holds. Such an unintuitive
inexecutability is not in $\INEXE{}{}$ and thus indicates the missing indirect
dependence.

  The general case is nevertheless more complex, and it seems that such
indirect dependences cannot be computed automatically in the case of
indeterminate effects (\cf\ the example in~\cite{CastilhoEtAl02}). We are
currently investigating this issue.

\myskip

  A different viewpoint of the work we presented here can be found
in~\cite{HerzigVarzinczak-IJCAI05}, where modularity of action theories is
assessed from a software engineering perspective. A modularity-based approach
for narrative reasoning about actions is given in~\cite{KakasEtAl05}.

\myskip

  Our postulates do not take into account causality statements linking
propositions such as those defined in~\cite{Lin95,McCainTurner95}. This could be
a topic for further investigation.

\section*{Acknowledgments}

  Ivan Varzinczak has been supported by a fellowship from the government of the
{\sc Federative Republic of Brazil}. Grant: CAPES BEX 1389/01-7.

\appendix

\section{Proof of Theorem~\ref{th_imp_stat}}\label{proof_th_imp_stat}

{\em An action theory $\ActTheory{\act}{}$ with $\depbin$ satisfies
Postulate~\PostulStatLaw\ if and only if $\IMPSTATtotal=\emptyset$.}

  We recall that $\models$ is logical consequence in Classical Propositional
Logic, and $\PI{\Set}$ is the set of prime implicates of the set \Set\ of
classical formulas.

  Before giving the proof of the theorem, we recall some properties of prime
implicates~\cite{Marquis95,Marquis2000} and the function
\NewCons{}{}~\cite{Inoue92}. Let $\fml\in\Fml$, $\Set\subseteq\Fml$, and
$\clause$ be a clause. Then
\begin{enumerate}
\item $\models\fml\sii\bigwedge\PI{\fml}$~\cite[Corollary
3.2]{Marquis2000}.\label{property-1}
\item $\PI{\Set}\cup\NewCons{\Set}{\fml}=\PI{\Set\et\fml}$ (from the definition
of \NewCons{}{}).\label{property-2}
\item $\models\Set\et\fml\sii\Set\et\NewCons{\Set}{\fml}$ (from~\ref{property-1}
and~\ref{property-2})\label{property-3}
\item If $\PI{\fml}\models\clause$, then there exists $\clause'\in\PI{\fml}$
such that $\clause'\models\clause$~\cite[Proposition 3.4]{Marquis2000}.
\label{property-4}
\end{enumerate}

Let $\depbin\subseteq\Act\times\Lit$, $\antec\imp\poss\act\top\in\EXE{\act}{}$,
$\CONSEQ{\act}{}=\EFF{\act}{}\cup\INEXE{\act}{}$, and $\hat{\CONSEQ{\act}{}}
\subseteq\CONSEQ{\act}{}$. We define:
\[
\antec_{\hat{\CONSEQ{\act}{}}}=\bigwedge\{\antec_{i}: \antec_{i}\imp\nec\act
\conseq_{i} \in \hat{\CONSEQ{\act}{}}\}
\]\[
\conseq_{\hat{\CONSEQ{\act}{}}}=\bigwedge\{\conseq_{i}: \antec_{i}\imp\nec\act
\conseq_{i} \in \hat{\CONSEQ{\act}{}}\}
\]


\begin{lemma}\label{lemma1}
$\STAT{}{}\cup\{\conseq_{\hat{\CONSEQ{\act}{}}}\}\cup\{\bigwedge_{\depbinvalid
\neg\lit_{j}\imp\nec\act\neg\lit_{j}}\neg\lit_{j}\}\Cvalid{}\bot$ if and only if
$\STAT{}{}\cup\NewCons{\STAT{}{}}{\conseq_{\hat{\CONSEQ{\act}{}}}}\cup\{
\bigwedge_{\depbinvalid\neg\lit_{j}\imp\nec\act\neg\lit_{j}}\neg\lit_{j}\}
\Cvalid{}\bot$.
\end{lemma}
\begin{proof}
Consequence of Property~\ref{property-3}.
\end{proof}

\begin{lemma}\label{lemma2}
If $\STAT{}{}\cup\NewCons{\STAT{}{}}{\conseq_{\hat{\CONSEQ{\act}{}}}}\cup
\{\bigwedge_{\depbinvalid\neg\lit_{j}\imp\nec\act\neg\lit_{j}}\neg\lit_{j}\}
\Cvalid{}\bot$, then $\exists\clause\in\NewCons{\STAT{}{}}{\conseq_{\hat{
\CONSEQ{\act}{}}}}$ such that $\STAT{}{}\cup\{\clause\}\cup\{\bigwedge_{
\depbinvalid\neg\lit_{j}\imp\nec\act\neg\lit_{j}}\neg\lit_{j}\}\Cvalid{}\bot$.
\end{lemma}
\begin{proof}
Consequence of Properties~\ref{property-1}, \ref{property-2}
and~\ref{property-4}.
\end{proof}

\begin{lemma}\label{lemma2b}
If $\STAT{}{}\cup\{\antec,\antec_{\hat{\CONSEQ{\act}{}}}\}\cup\{\bigwedge_{
\depbinvalid\neg\lit_{j}\imp\nec\act\neg\lit_{j}}\neg\lit_{j}\}\notCvalid{}\bot$
and $\STAT{}{}\cup\NewCons{\STAT{}{}}{\conseq_{\hat{\CONSEQ{\act}{}}}}\cup
\{\bigwedge_{\depbinvalid\neg\lit_{j}\imp\nec\act\neg\lit_{j}}\neg\lit_{j}\}
\Cvalid{}\bot$, then $\exists\clause\in\NewCons{\STAT{}{}}{\conseq_{\hat{\CONSEQ
{\act}{}}}}$ such that $\STAT{}{}\cup\{\clause\}\cup\{\bigwedge_{\depbinvalid
\neg\lit_{j}\imp\nec\act\neg\lit_{j}}\neg\lit_{j}\}\Cvalid{}\bot$.
\end{lemma}
\begin{proof}
By Lemma~\ref{lemma2} and Classical Logic.
\end{proof}

\begin{lemma}\label{lemma3}
If $\STAT{}{}\cup\{\antec,\antec_{\hat{\CONSEQ{\act}{}}}\}\cup\{\bigwedge_{
\depbinvalid\neg\lit_{j}\imp\nec\act\neg\lit_{j}}\neg\lit_{j}\}\notCvalid{}\bot$
and $\STAT{}{}\cup\NewCons{\STAT{}{}}{\conseq_{\hat{\CONSEQ{\act}{}}}}\cup
\{\bigwedge_{\depbinvalid\neg\lit_{j}\imp\nec\act\neg\lit_{j}}\neg\lit_{j}\}
\Cvalid{}\bot$, then $\exists\clause\in\NewCons{\STAT{}{}}{\conseq_{\hat{
\CONSEQ{\act}{}}}}$ such that both $\STAT{}{}\cup\{\antec,\antec_{\hat{\CONSEQ{
\act}{}}}\}\cup\{\bigwedge_{\depbinvalid\neg\lit_{j}\imp\nec\act\neg\lit_{j}}
\neg\lit_{j}\}\notCvalid{}\bot$ and $\STAT{}{}\cup\{\clause\}\cup\{\bigwedge_{
\depbinvalid\neg\lit_{j}\imp\nec\act\neg\lit_{j}}\neg\lit_{j}\}\Cvalid{}\bot$.
\end{lemma}
\begin{proof}
Trivially, by Lemma~\ref{lemma2b}. 
\end{proof}

\bigskip

\begin{lemma}\label{lemma4}
If $\clause\in\NewCons{\STAT{}{}}{\conseq_{\hat{\CONSEQ{\act}{}}}}$ is such
that $\STAT{}{}\cup\{\antec,\antec_{\hat{\CONSEQ{\act}{}}}\}\cup\{\bigwedge_{
\depbinvalid\neg\lit_{j}\imp\nec\act\neg\lit_{j}}\neg\lit_{j}\}\notCvalid{}\bot$
and $\STAT{}{}\cup\{\clause\}\cup\{\bigwedge_{\depbinvalid\neg\lit_{j}\imp\nec
\act\neg\lit_{j}}\neg\lit_{j}\}\Cvalid{}\bot$, then $\STAT{}{}\cup\{\antec,
\antec_{\hat{\CONSEQ{\act}{}}}\}\cup\{\bigwedge_{\stackrel{\lit_{i}\in\clause}
{\act\indepbin\lit_{i}}}\neg\lit_{i}\}\notCvalid{}\bot$ and $\STAT{}{}\cup
\{\clause\}\cup\{\bigwedge_{\stackrel{\lit_{i}\in\clause}{\act\indepbin\lit_{i}}}
\neg\lit_{i}\}\Cvalid{}\bot$.
\end{lemma}
\begin{proof}
Let $\clause\in\NewCons{\STAT{}{}}{\conseq_{\hat{\CONSEQ{\act}{}}}}$ be such 
that $\STAT{}{}\cup\{\antec,\antec_{\hat{\CONSEQ{\act}{}}}\}\cup\{\bigwedge_{
\depbinvalid\neg\lit_{j}\imp\nec\act\neg\lit_{j}}\neg\lit_{j}\}\notCvalid{}\bot$
and $\STAT{}{}\cup\{\clause\}\cup\{\bigwedge_{\depbinvalid\neg\lit_{j}\imp\nec
\act\neg\lit_{j}}\neg\lit_{j}\}\Cvalid{}\bot$.

If $\clause=\bot$, the result is trivial.

Let $\atm{\fml}$ denote the set of atoms occurring in a classical formula
$\fml$.
\begin{itemize}
\item If $\atm{\clause}\not\subset\atm{\bigwedge_{\depbinvalid\neg\lit_{j}\imp
\nec\act\neg\lit_{j}}\neg\lit_{j}}$, then the premise is false (and the lemma
trivially holds).
\item If $\atm{\clause}=\atm{\bigwedge_{\depbinvalid\neg\lit_{j}\imp\nec\act\neg
\lit_{j}}\neg\lit_{j}}$, the lemma holds.
\item Let $\atm{\clause}\subset\atm{\bigwedge_{\depbinvalid\neg\lit_{j}\imp\nec
\act\neg\lit_{j}}\neg\lit_{j}}$. From $\STAT{}{}\cup\{\antec,\antec_{\hat{
\CONSEQ{\act}{}}}\}\cup\{\bigwedge_{\depbinvalid\neg\lit_{j}\imp\nec\act\neg
\lit_{j}}\neg\lit_{j}\}\notCvalid{}\bot$ it follows $\STAT{}{}\cup\{\antec,
\antec_{\hat{\CONSEQ{\act}{}}}\}\cup\{\bigwedge_{\stackrel{\lit_{i}\in\clause}
{\act\indepbin\lit_{i}}}\neg\lit_{i}\}\notCvalid{}\bot$. From $\STAT{}{}\cup
\{\clause\}\cup\{\bigwedge_{\depbinvalid\neg\lit_{j}\imp\nec\act\neg\lit_{j}}
\neg\lit_{j}\}\Cvalid{}\bot$ and because $\STAT{}{}\cup\{\bigwedge_{\depbinvalid
\neg\lit_{j}\imp\nec\act\neg\lit_{j}}\neg\lit_{j}\}\notCvalid{}\bot$, it follows
$\STAT{}{}\cup\{\clause\}\cup\{\bigwedge_{\stackrel{\lit_{i}\in\clause}{\act
\indepbin\lit_{i}}}\neg\lit_{i}\}\Cvalid{}\bot$.
\end{itemize}
\end{proof}

\begin{lemma}\label{lemma5}
If $\clause\in\NewCons{\STAT{}{}}{\conseq_{\hat{\CONSEQ{\act}{}}}}$ is such that
$\STAT{}{}\cup\{\antec,\antec_{\hat{\CONSEQ{\act}{}}}\}\cup\{\bigwedge_{
\stackrel{\lit_{i}\in\clause}{\act\indepbin\lit_{i}}}\neg\lit_{i}\}\notCvalid{}
\bot$ and $\STAT{}{}\cup\{\clause\}\cup\{\bigwedge_{\stackrel{\lit_{i}\in
\clause}{\act\indepbin\lit_{i}}}\neg\lit_{i}\}\Cvalid{}\bot$, then $\STAT{}{}
\cup\{\antec,\antec_{\hat{\CONSEQ{\act}{}}}\}\cup\{\bigwedge_{\stackrel{\lit_{i}
\in\clause}{\act\indepbin\lit_{i}}}\neg\lit_{i}\}\notCvalid{}\bot$ and $\forall
\lit_{i}\in\clause,\ \act\indepbin\lit_{i}$.
\end{lemma}
\begin{proof}
From $\STAT{}{}\cup\{\antec,\antec_{\hat{\CONSEQ{\act}{}}}\}\cup\{\bigwedge_{
\stackrel{\lit_{i}\in\clause}{\act\indepbin\lit_{i}}}\neg\lit_{i}\}\notCvalid{}
\bot$ we conclude $\STAT{}{}\cup\{\bigwedge_{\stackrel{\lit_{i}\in\clause}{\act
\indepbin\lit_{i}}}\neg\lit_{i}\}\notCvalid{}\bot$. From this and the hypothesis
$\STAT{}{}\cup\{\clause\}\cup\{\bigwedge_{\stackrel{\lit_{i}\in\clause}{\act
\indepbin\lit_{i}}}\neg\lit_{i}\}\Cvalid{}\bot$, it follows $\STAT{}{}\cup
\{\bigwedge_{\stackrel{\lit_{i}\in\clause}{\act\indepbin\lit_{i}}}\neg\lit_{i}\}
\Cvalid{}\neg\clause$. If $\STAT{}{}\models\neg\clause$, then 
$\STAT{}{},{\conseq_{\hat{\CONSEQ{\act}{}}}}\models\neg\clause$, and because
$\clause\in\NewCons{\STAT{}{}}{\conseq_{\hat{\CONSEQ{\act}{}}}}$, we have
$\clause\models\neg\clause$, a contradiction. Hence
$\STAT{}{}\cup\{\clause\}\not\models\bot$. Suppose now that there is at least
one literal $\lit\in\clause$ such that $\neg\lit$ does not appear in
$\bigwedge_{\stackrel{\lit_{i}\in\clause}{\act\indepbin\lit_{i}}}\neg\lit_{i}$.
Then, the propositional valuation in which $\clause_{\lit\leftarrow\verum}$
satisfies $\STAT{}{}\cup\{\clause\}\cup\bigwedge_{\stackrel{\lit_{i}\in\clause}
{\act\indepbin\lit_{i}}}\neg\lit_{i}$, and then $\STAT{}{},\{\clause\},
\bigwedge_{\stackrel{\lit_{i}\in\clause}{\act\indepbin\lit_{i}}}\neg\lit_{i}\not
\models\bot$. Hence there cannot be such a literal, and then $\forall\lit_{i}\in
\clause,\ \act\indepbin\lit_{i}$.
\end{proof}

\myskip

{\bf Proof of Theorem~\ref{th_imp_stat}}

\noindent($\Rightarrow$): Suppose $\IMPSTATtotal\neq\emptyset$. Then at the
first step of the algorithm there has been some $\antec\imp\poss\act\top\in
\EXE{\act}{}$ and some $\hat{\CONSEQ{\act}{}}\subseteq\CONSEQ{\act}{}$ such that
$\STAT{}{},\EFF{\act}{},\EXE{\act}{},\INEXE{\act}{}\depbinvalid\neg(\antec\et
\antec_{\hat{\CONSEQ{\act}{}}})$ and $\STAT{}{}\not\models\neg(\antec\et\antec_
{\hat{\CONSEQ{\act}{}}})$. Hence $\ActTheory{\act}{}$ with $\depbin$ does not
satisfy Postulate~\PostulStatLaw.

\noindent($\Leftarrow$): Suppose that $\IMPSTATtotal=\emptyset$. Therefore for
all $\antec'\imp\poss\act\top\in\EXE{\act}{}$ and for all subsets
$\hat{\CONSEQ{\act}{}}\subseteq\CONSEQ{\act}{}$, we have that\\
\begin{equation}\label{negation_alg_test}
\text{$\forall\clause\in\NewCons{\STAT{}{}}{\conseq_{\hat{\CONSEQ{\act}{}}}}$
              if $\STAT{}{}\cup\{\antec',\antec_{\hat{\CONSEQ{\act}{}}}, 
              \neg\clause\}\notCvalid{}\bot$,
              then $\exists\lit_{i}\in\clause,\ \act\depbin\lit_{i}$}
\end{equation}

From~(\ref{negation_alg_test}), the contraposition
of~Lemmas~\ref{lemma5}--\ref{lemma2b}, and~Lemma~\ref{lemma1}, it follows that
for all $\antec'\imp\poss\act\top\in\EXE{\act}{}$ and $\hat{\CONSEQ{\act}{}}
\subseteq\CONSEQ{\act}{}$,

\begin{equation}\label{hypothesis}
\text{if $\STAT{}{}\cup\{\antec',\antec_{\hat{\CONSEQ{\act}{}}}\}\cup
    \{\bigwedge_{\depbinvalid\neg\lit_{j}\imp\nec\act\neg\lit_{j}}\neg\lit_{j}\}
    \notCvalid{}\bot$, then  
    $\STAT{}{}\cup\{\conseq_{\hat{\CONSEQ{\act}{}}}\}\cup\{\bigwedge_{
     \depbinvalid\neg\lit_{j}\imp\nec\act\neg\lit_{j}}\neg\lit_{j}\}
     \notCvalid{}\bot$.} 
\end{equation}

  Now, suppose $\STAT{}{}\not\models\fml$ for some propositional $\fml$. We will
build a model $\model$ such that $\model$ is a $\depbin$-model for
$\ActTheory{\act}{}$ that does not satisfy $\fml$. Let $\Worlds$ be the set of
all propositional valuations satisfying $\STAT{}{}$ that falsify $\fml$. As
$\STAT{}{}\not\models\fml$, $\STAT{}{}\cup\{\neg\fml\}$ is satisfiable, hence
$\Worlds$ must be nonempty. 
We define the binary relation $\ract$ on $\Worlds$ such that $w\ract w'$ if and
only if for every $\antec\imp\nec\act\conseq\in\CONSEQ{\act}{}$ such that
$\wMvalid{w}{\model}\antec$:
\begin{itemize}
\item $\wMvalid{w'}{\model}\conseq$; and
\item $\wMvalid{w'}{\model}\neg\lit_{j}$ for all $\lit_{j}$ such that
$\act\indepbin\lit_{j}$ and $\wMvalid{w}{\model}\neg\lit_{j}$.
\end{itemize}
Taking the obvious definition of $\Val$ we obtain a model $\model=\PDLmodel$. We
have that $\model$ is a $\depbin$-model, by the definition of $\ract$, and
$\Mvalid{\model}\STAT{}{}\et\EFF{\act}{}\et\EXE{\act}{}\et\INEXE{\act}{}$,
because:
\begin{itemize}
\item $\Mvalid{\model}\STAT{}{}$: by definition of $\Worlds$;
\item $\Mvalid{\model}\CONSEQ{\act}{}$: for every world $w$ and every 
$\antec\imp\nec\act\conseq\in\CONSEQ{\act}{}$, if $\wMvalid{w}{\model}\antec$,
then, by the definition of $\ract$, $\wMvalid{w'}{\model}\conseq$ for all
$w'\in\Worlds$ such that $w\ract w'$;
\item $\Mvalid{\model}\EXE{\act}{}$: let $\EFF{\act}{}(w)=\{\antec\imp\nec\act
\conseq\in\EFF{\act}{}: \wMvalid{w}{\model}\antec\}$, and $\indep{\act}{w}=
\{\neg\lit: \act\indepbin\lit \text{ and }\wMvalid{w}{\model}\neg\lit\}$. Then,
for every world $w$ and every $\antec'\imp\poss\act\top\in\EXE{\act}{}$, if
$\wMvalid{w}{\model}\antec'\et\antec_{\EFF{\act}{}(w)}\et\indep{\act}{w}$, then
from~(\ref{hypothesis}), $\conseq_{\EFF{\act}{}(w)}\et\indep{\act}{w}\not\models
\bot$. As $\Worlds$ is maximal, there exists at least one $w'$ such that
$\wMvalid{w'}{\model}\conseq_{\EFF{\act}{}(w)}\et\indep{\act}{w}$. As $\ract$ is
maximal by definition, we have $w\ract w'$.

and the definition of $\ract$, there exists at least
one $w'$ such that $w\ract w'$.
\end{itemize}
Clearly $\notMvalid{\model}\fml$, by the definition of $\Worlds$. Hence
$\STAT{}{},\EFF{\act}{},\EXE{\act}{},\INEXE{\act}{}\notdepbinvalid\fml$.
Therefore $\ActTheory{\act}{}$ and $\depbin$ violate
Postulate~\PostulStatLaw.{\hfill \qed}

\section{Proof of Theorem~\ref{th_imp_inexe}}\label{proof_th_imp_inexe}

{\em If $\ActTheory{\act}{}$ with $\depbin$ satisfies Postulate~\PostulStatLaw,
then it satisfies Postulate~\PostulInexeLaw\ if and only if $\IMPINEXE{\act}{}=
\emptyset$.}

Let $\depbin\subseteq\Act\times\Lit$ and
$\antec\imp\poss\act\top\in\EXE{\act}{}$. For every
$\hat{\EFF{\act}{}}\subseteq\EFF{\act}{}$ we define:
\[
\antec_{\hat{\EFF{\act}{}}}=\bigwedge\{\antec_{i}: \antec_{i}\imp\nec\act
\conseq_{i} \in \hat{\EFF{\act}{}}\}
\]\[
\conseq_{\hat{\EFF{\act}{}}}=\bigwedge\{\conseq_{i}: \antec_{i}\imp\nec\act
\conseq_{i} \in \hat{\EFF{\act}{}}\}
\]Moreover, we define
\[
\IMPINEXE{\act}{}=\{(\antec_{\hat{\EFF{\act}{}}}\et\neg\clause)\imp\nec\act\bot:
\hat{\EFF{\act}{}}\subseteq\EFF{\act}{}, \STAT{}{}\cup\{\antec_{\hat{\EFF{
\act}{}}}, \neg\clause\}\not\models\bot,\clause\in\NewCons{\STAT{}{}}
{\conseq_{\hat{\EFF{\act}{}}}},\act\indepbin\lit_{i}\forall\lit_{i}\in\clause\}
\]

\begin{lemma}\label{l0}
If $\ActTheory{\act}{}$ satisfies Postulate~\PostulStatLaw, then 
$\STAT{}{},\EFF{\act}{},\EXE{\act}{},\INEXE{\act}{}\depbinvalid
\antec\imp\nec\act\bot$ implies $\STAT{}{},\EFF{\act}{},\INEXE{\act}{}
\depbinvalid\antec\imp\nec\act\bot$.
\end{lemma}
\begin{proof}
Straightforward as a special case of Theorem~\ref{th_intra_mod_eff}.
\end{proof}

\begin{lemma}\label{l1}
If for each $\hat{\EFF{\act}{}}$ we have
$\STAT{}{}\cup\{\antec_{\hat{\EFF{\act}{}}}\}\not\models\bot$ implies 
$\STAT{}{}\cup\{\conseq_{\hat{\EFF{\act}{}}}\}\not\models\bot$, then
$\STAT{}{},\EFF{\act}{},\INEXE{\act}{}\PDLvalid\antec\imp\nec\act\bot$ implies
$\STAT{}{},\INEXE{\act}{}\PDLvalid\antec\imp\nec\act\bot$.
\end{lemma}
\begin{proof}
If $\STAT{}{},\INEXE{\act}{}\notPDLvalid\antec\imp\nec\act\bot$, then there
exists a \PDL-model $\model=\PDLmodel$ such that
$\Mvalid{\model}\STAT{}{}\et\INEXE{\act}{}$, and there is a possible world 
$w\in\Worlds$ such that $\wMvalid{w}{\model}\antec$ and
$\wMvalid{w}{\model}\poss\act\top$.

(We are going to construct a counter-model for
$\STAT{}{},\EFF{\act}{},\INEXE{\act}{}\PDLvalid\antec\imp\nec\act\bot$.)

Let $\EFF{\act}{}(w)=\{\antec\imp\nec\act\conseq\in\EFF{\act}{}:
\wMvalid{w}{\model}\antec\}$. Then $\antec_{\EFF{\act}{}(w)}=
\bigwedge\{\antec_{i}:\antec_{i}\imp\nec\act\conseq_{i}\in\EFF{\act}{}(w)\}$ is
such that $\wMvalid{w}{\model}\antec_{\EFF{\act}{}(w)}$. As $\STAT{}{}\et
\antec_{\EFF{\act}{}(w)}$ is thus satisfiable, $\STAT{}{}\et
\conseq_{\EFF{\act}{}(w)}$, with $\conseq_{\EFF{\act}{}(w)}=
\bigwedge\{\conseq_{i}:\antec_{i}\imp\nec\act\conseq_{i}\in\EFF{\act}{}(w)\}$,
must be satisfiable, too (by hypothesis, because $\EFF{\act}{}(w)\subseteq
\EFF{\act}{}$). Hence, there exists a propositional valuation \val\ such that
$\val(\STAT{}{}\et\conseq_{\EFF{\act}{}(w)})=\true$.

Consider, thus, $v$ such that $v\notin\Worlds$, and extend $\Val$ such that
$\Val(v)=\val$. Let $\model'=\tuple{\Worlds',\AccRel',\Val'}$ be such that 
$\Worlds'=\Worlds\cup\{v\}$, $\AccRel'_{\act}(w)=\{v\}$ for all $u$ such that
$\wMvalid{u}{\model}\antec$ and $\AccRel'_{\act}(u)=\emptyset$ otherwise,
and $\Val'=\Val\cup\{(v,\Val(v))\}$.

Then:
\begin{itemize}
\item $\Mvalid{\model'}\STAT{}{}$ because $\Mvalid{\model}\STAT{}{}$ and
$\val(\STAT{}{})=\true$.
\item $\wMvalid{v}{\model'}\INEXE{\act}{}$ because
$\wMvalid{v}{\model'}\nec\act\bot$, by definition of $\AccRel'_{\act}(v)$, and
$\wMvalid{w}{\model'}\INEXE{\act}{}$ because $\notwMvalid{w}{\model'}\antec$ for
all $\antec\imp\nec\act\bot\in\INEXE{\act}{}$, as $\notwMvalid{w}{\model}
\antec$ (otherwise, as $\Mvalid{\model}\INEXE{\act}{}$, we would not have
$\wMvalid{w}{\model}\poss\act\top$).
\item $\wMvalid{v}{\model'}\EFF{\act}{}$ because $\wMvalid{v}{\model'}\nec\act
\bot$, and $\wMvalid{w}{\model'}\EFF{\act}{}$ by construction of $\model'$:
$\AccRel'_{\act}(w)=\{v\}$ and $\wMvalid{v}{\model'}\conseq_{\EFF{\act}{}(w)}$.
\item $\wMvalid{w}{\model'}\antec\et\poss\act\top$.
\end{itemize}

Hence $\model'$ is still a model of $\STAT{}{}$, $\INEXE{\act}{}$ and
$\EFF{\act}{}$. Of course, $\model'$ is a counter-model for
$\antec\imp\nec\act\bot$.
\end{proof}

\begin{lemma}\label{l2}
Let $\FRA{\act}{}=\{\neg\lit\imp\nec\act\neg\lit: \act\indepbin\lit\}$. Then if
$\STAT{}{},\EFF{\act}{},\INEXE{\act}{},\IMPINEXE{\act}{},\FRA{\act}{}\PDLvalid
\antec\imp\nec\act\bot$, then $\STAT{}{},\EFF{\act}{},\INEXE{\act}{},
\IMPINEXE{\act}{}\PDLvalid\antec\imp\nec\act\bot$.
\end{lemma}
\begin{proof}
If $\STAT{}{},\EFF{\act}{},\INEXE{\act}{},\IMPINEXE{\act}{}\notPDLvalid
\antec\imp\nec\act\bot$, then there exists a \PDL-model $\model=\PDLmodel$ such
that $\Mvalid{\model}\STAT{}{}\et\EFF{\act}{}\et\INEXE{\act}{}\et
\IMPINEXE{\act}{}$, and there is a possible world $w\in\Worlds$ such that
$\wMvalid{w}{\model}\antec$ and $\wMvalid{w}{\model}\poss\act\top$.

We are going to construct a counter-model for $\STAT{}{},\EFF{\act}{},
\INEXE{\act}{},\IMPINEXE{\act}{},\FRA{\act}{}\PDLvalid\antec\imp\nec\act\bot$.
Let $\model'=\tuple{\Worlds',\AccRel',\Val'}$ be such that $\Worlds'=\Worlds$,
and $\AccRel'_{\act}(w')=\emptyset$ for every $w'\neq w$, and $\Val'=\Val$.

Of course, $\model'$ is still a model of $\STAT{}{}$, $\EFF{\act}{}$,
$\INEXE{\act}{}$ and $\IMPINEXE{\act}{}$.

For every $\hat{\EFF{\act}{}}\subseteq\EFF{\act}{}$, the case where
$\wMvalid{w}{\model'}\antec_{\hat{\EFF{\act}{}}}\et\neg\clause$, with
$\clause\in\NewCons{\STAT{}{}}{\conseq_{\hat{\EFF{\act}{}}}}$,
$\act\indepbin\lit_{i},\forall\lit_{i}\in\clause$, is impossible, because
$\Mvalid{\model}\IMPINEXE{\act}{}$ and hence we would have
$\AccRel_{\act}(w)=\emptyset$, contradicting the hypothesis that
$\wMvalid{w}{\model}\poss\act\top$.

Thus, we have to consider only the following cases:
\begin{itemize}
\item if $\wMvalid{w}{\model'}\lit_{i}$, for every $\lit_{i}$ such that $\act
\indepbin\lit_{i}$, then $\model'$ is also a model of $\FRA{\act}{}$, and then
we have a counter-model for $\STAT{}{},\EFF{\act}{},\INEXE{\act}{},
\IMPINEXE{\act}{},\FRA{\act}{}\PDLvalid\antec\imp\nec\act\bot$.
\item if $\wMvalid{w}{\model'}\antec_{\hat{\EFF{\act}{}}}\et\bigwedge\neg
\lit_{i}$, where $\act\indepbin\lit_{i}$ and there is no clause $\clause\in
\NewCons{\STAT{}{}}{\conseq_{\hat{\EFF{\act}{}}}}$ such that $\lit_{i}\in\clause$,
for some $\hat{\EFF{\act}{}}\subseteq\EFF{\act}{}$, then of course 
$\conseq_{\hat{\EFF{\act}{}}}\et\bigwedge\neg\lit_{i}$ is satisfiable, \ie,
there is a valuation where
$\conseq_{\hat{\EFF{\act}{}}}\et\bigwedge\neg\lit_{i}$ holds. Let
$\val_{\hat{\EFF{\act}{}}}$ be such a valuation.

Consider, thus, $v$ such that $v\notin\Worlds$, and extend $\Val'$ such that
$\Val'(v)=\val_{\hat{\EFF{\act}{}}}$. Now let
$\model''=\tuple{\Worlds'',\AccRel'', \Val''}$ be such that
$\Worlds''=\Worlds'\cup\{v\}$, and
$\AccRel''_{\act}(w)=\val_{\hat{\EFF{\act}{}}}$, and
$\Val''=\Val'\cup\{(v,\Val'(v))\}$.

Again, it can easily be checked that $\model''$ is a model of $\STAT{}{}$,
$\EFF{\act}{}$, $\INEXE{\act}{}$ and $\IMPINEXE{\act}{}$. Moreover, it is a
model of $\FRA{\act}{}$, and hence a model for $\STAT{}{},\EFF{\act}{},\INEXE{
\act}{},\IMPINEXE{\act}{},\FRA{\act}{}$ and $\antec\et\poss\act\top$.
\end{itemize}
\end{proof}

\begin{lemma}\label{l3}
If $\STAT{}{},\EFF{\act}{},\INEXE{\act}{}\depbinvalid\antec\imp\nec\act\bot$, then
$\STAT{}{},\INEXE{\act}{},\IMPINEXE{\act}{}\PDLvalid\antec\imp\nec\act\bot$.
\end{lemma}
\begin{proof}
Let
\[
\EFFp{\act}{}=\{\antec_{\hat{\EFF{\act}{}}}\imp\nec\act\conseq_{\hat{\EFF{
\act}{}}}: \hat{\EFF{\act}{}}\subseteq\EFF{\act}{}, \STAT{}{}\cup\{\antec_{\hat{
\EFF{\act}{}}}\}\not\models\bot, \STAT{}{}\cup\{\conseq_{\hat{\EFF{\act}{}}}\}
\not\models\bot\}
\]\[
\EFFm{\act}{}=\{\antec_{\hat{\EFF{\act}{}}}\imp\nec\act\conseq_{\hat{\EFF{
\act}{}}}: \hat{\EFF{\act}{}}\subseteq\EFF{\act}{}, \STAT{}{}\cup\{\antec_{\hat{
\EFF{\act}{}}}\}\not\models\bot, \STAT{}{}\models\conseq_{\hat{\EFF{\act}{}}}
\imp\bot\}
\]
The following steps establish the result.

\begin{list}
{\arabic{proof_item}.}{\usecounter{proof_item}}
\item $\STAT{}{},\EFF{\act}{},\INEXE{\act}{}\depbinvalid\antec\imp\nec\act\bot$,
by hypothesis\label{it1}
\item $\STAT{}{},\EFF{\act}{},\INEXE{\act}{},\IMPINEXE{\act}{}\depbinvalid\antec
\imp\nec\act\bot$, from~\ref{it1}.\ and monotonicity\label{it4}
\item $\STAT{}{}\PDLvalid\bigwedge\EFF{\act}{}\sii(\bigwedge\EFFp{\act}{}\et
\bigwedge\EFFm{\act}{})$, by definition of $\EFFp{\act}{}$ and $\EFFm{\act}{}$,
and \PDL\label{it5}
\item $\STAT{}{},\EFFp{\act}{}\cup\EFFm{\act}{},\INEXE{\act}{},\IMPINEXE{\act}{}
\depbinvalid\antec\imp\nec\act\bot$, from~\ref{it4}.\ and~\ref{it5}.\label{it6}
\item $\EFFm{\act}{}\subseteq\IMPINEXE{\act}{}$, as if $\antec_{\hat{\EFF{
\act}{}}}\imp\nec\act\conseq_{\hat{\EFF{\act}{}}}\in\EFFm{\act}{}$, then
$\STAT{}{}\models\conseq_{\hat{\EFF{\act}{}}}\imp\bot$, and then
$\bot\in\NewCons{\STAT{}{}}{\conseq_{\hat{\EFF{\act}{}}}}$, from which it follows
that 
$(\antec_{\hat{\EFF{\act}{}}}\et\top)\imp\nec\act\bot\in\IMPINEXE{\act}{}$, and
then $\antec_{\hat{\EFF{\act}{}}}\imp\nec\act\bot\in\IMPINEXE{\act}{}$\label{it7}
\item $\STAT{}{},\EFFp{\act}{},\INEXE{\act}{},\IMPINEXE{\act}{}\depbinvalid
\antec\imp\nec\act\bot$, from~\ref{it6}.\ and~\ref{it7}.\label{it8}
\item $\STAT{}{},\EFFp{\act}{},\INEXE{\act}{},\IMPINEXE{\act}{},\FRA{\act}{}
\PDLvalid\antec\imp\nec\act\bot$, from~\ref{it8}.\ and definition of $\depbin$,
where $\FRA{\act}{}=\{\neg\lit\imp\nec\act\neg\lit: \act\indepbin\lit\}$.
\label{it9}
\item $\STAT{}{},\EFFp{\act}{},\INEXE{\act}{},\IMPINEXE{\act}{}\PDLvalid
\antec\imp\nec\act\bot$, from~\ref{it9}.\ and~Lemma~\ref{l2}.\label{it10}
\item $\STAT{}{},\INEXE{\act}{},\IMPINEXE{\act}{}\PDLvalid\antec\imp\nec\act\bot$,
from~\ref{it10}.\ and~Lemma~\ref{l1}, whose hypothesis is satisfied by the
definition of $\EFFp{\act}{}$.
\end{list}
\end{proof}

{\bf Proof of Theorem~\ref{th_imp_inexe}}


\noindent($\Rightarrow$): Straightforward, as every time
$\STAT{}{},\EFF{\act}{},\EXE{\act}{},\INEXE{\act}{}\depbinvalid
\antec\imp\nec\act\bot$, we have 
$\STAT{}{},\INEXE{\act}{}\PDLvalid\antec\imp\nec\act\bot$, and then
$\IMPINEXE{\act}{}$ never changes.

\bigskip

\noindent($\Leftarrow$): We are going to show that if
$\STAT{}{},\EFF{\act}{},\EXE{\act}{},\INEXE{\act}{}\depbinvalid
\antec\imp\nec\act\bot$ and $\IMPINEXE{\act}{}=\emptyset$, then
$\STAT{}{},\INEXE{\act}{}\PDLvalid\antec\imp\nec\act\bot$.

\begin{list}
{\arabic{proof_item}.}{\usecounter{proof_item}\setcounter{proof_item}{0}}
\item $\STAT{}{},\EFF{\act}{},\EXE{\act}{},\INEXE{\act}{}\depbinvalid
\antec\imp\nec\act\bot$, by hypothesis\label{ita}
\item $\STAT{}{},\EFF{\act}{},\INEXE{\act}{}\depbinvalid
\antec\imp\nec\act\bot$, from~\ref{ita}.\ and Lemma~\ref{l0}\label{itb}
\item $\STAT{}{},\INEXE{\act}{},\IMPINEXE{\act}{}\PDLvalid\antec\imp\nec\act\bot$,
from~\ref{itb}.\ and Lemma~\ref{l3}\label{itc}
\item $\STAT{}{},\INEXE{\act}{}\PDLvalid\antec\imp\nec\act\bot$, from~\ref{itc}.\
and hypothesis $\IMPINEXE{\act}{}=\emptyset$.
\end{list}{\hfill \qed}


\end{document}